\documentclass[letterpaper]{article} 
\usepackage{aaai2027}  
\usepackage[hyphens]{url}  
\usepackage{graphicx} 
\usepackage{natbib}  
\usepackage{caption} 
\usepackage{algorithm}
\usepackage{algorithmic}

\usepackage{newfloat}
\usepackage{listings}
\DeclareCaptionStyle{ruled}{labelfont=normalfont,labelsep=colon,strut=off} 
\floatstyle{ruled}
\newfloat{listing}{tb}{lst}{}
\floatname{listing}{Listing}

\usepackage{booktabs}

\newcommand{\cmark}{\textcolor{green!60!black}{\ding{51}}} 
\newcommand{\xmark}{\textcolor{red}{\ding{55}}}           

\usepackage{latexsym}
\usepackage{booktabs}
\usepackage{algorithm}
\usepackage{arydshln}
\usepackage{multicol}
\usepackage{amsfonts}
\usepackage{graphicx}
\usepackage{multirow}
\usepackage{xcolor}
\usepackage{makecell} 
\usepackage{array} 
\usepackage{fontawesome} 
\usepackage{xcolor} 
\usepackage{amssymb} 
\usepackage{amsmath} 
\usepackage{colortbl}
\usepackage[listings,skins,breakable]{tcolorbox}
\usepackage{pifont}
\usepackage{listings}
\usepackage{subfigure}
\usepackage{graphicx}
\usepackage{marvosym}

\definecolor{lightgray}{gray}{0.95}
\definecolor{deepblue}{RGB}{70,130,180}
\definecolor{deepgray}{RGB}{119,136,153}
\lstdefinestyle{prompt}{
    basicstyle=\ttfamily\fontsize{7pt}{8pt}\selectfont,
    frame=none,
    breaklines=true,
    backgroundcolor=\color{lightgray},
    breakatwhitespace=true,
    breakindent=0pt,
    escapeinside={(*@}{@*)},
    numbers=none,
    numbersep=5pt,
    xleftmargin=5pt,
    aboveskip=2pt,
    belowskip=2pt,
}
\tcbset{
  aibox/.style={
    top=10pt,
    colback=white,
    enhanced,
    center,
  }
}
\newtcolorbox{AIbox}[2][]{aibox, title=#2,#1}

\title{DeepSurvey-Bench: Evaluating Academic Value of Automatically Generated Scientific Surveys}
\author{
    Guo-Biao Zhang\textsuperscript{\rm 1}, Xian-Ling Mao\textsuperscript{\rm 1}\thanks{Corresponding author}, Ding-Yuan Liu\textsuperscript{\rm 1}, Da-Yi Wu\textsuperscript{\rm 1},\\ Tian Lan\textsuperscript{\rm 1}, Huihui Li\textsuperscript{\rm 2}, Heyan Huang\textsuperscript{\rm 1}
}
\affiliations{
   \textsuperscript{\rm 1}School of Computer Science and Technology, Beijing Institute of Technology, China \\
   \textsuperscript{\rm 2}Guangdong Polytechnic Normal University
School of Computer Science, Zhongshan Avenue, Guangzhou 510665, China\\

    \{zgb\_stubborn, maoxl, 3220251327, 3220251256, hhy\}@bit.edu.cn, lantiangmftby@gmail.com, lihh@gpnu.edu.cn
}

\begin{document}

\maketitle

\begin{abstract}
The rapid development of automated scientific survey generation technology has made it increasingly important to establish a comprehensive benchmark to evaluate the quality of generated surveys. Most existing benchmarks first construct ground-truth survey datasets by selecting human-written surveys based on limited selection criteria, such as citation counts and structural coherence, and then evaluate generated surveys primarily based on conventional quality dimensions, including structural quality and reference relevance.
However, these benchmarks have two key issues: 
(1) the datasets are insufficiently reliable because the selection criteria only identify highly cited or structurally coherent surveys without verifying their academic value;
(2) the evaluation metrics mainly reflect the surface-level quality of generated surveys and are insufficient to assess their academic value. 
Together, these issues prevent existing benchmarks from effectively assessing the academic value of generated surveys. To address the above problems, we propose \textbf{DeepSurvey-Bench}, a comprehensive benchmark for evaluating the academic value of automatically generated scientific surveys. Specifically, our proposed benchmark introduces a set of academic value evaluation criteria covering three dimensions: informational value, scholarly communication value, and research guidance value. We first construct a reliable survey dataset with academic value annotations based on these criteria, and then evaluate the academic value of generated surveys according to these criteria through a multi-LLM-as-a-judge approach. Extensive experiments demonstrate that DeepSurvey-Bench not only aligns closely with human assessments in evaluating the academic value of surveys, but also reveals underlying academic value beyond the reach of surface-level quality metrics, providing a foundation for fine-grained diagnosis and iterative improvement of generated surveys. 

\end{abstract}


\section{Introduction}
\label{intro}

\begin{figure}[t]
  \includegraphics[width=\columnwidth]{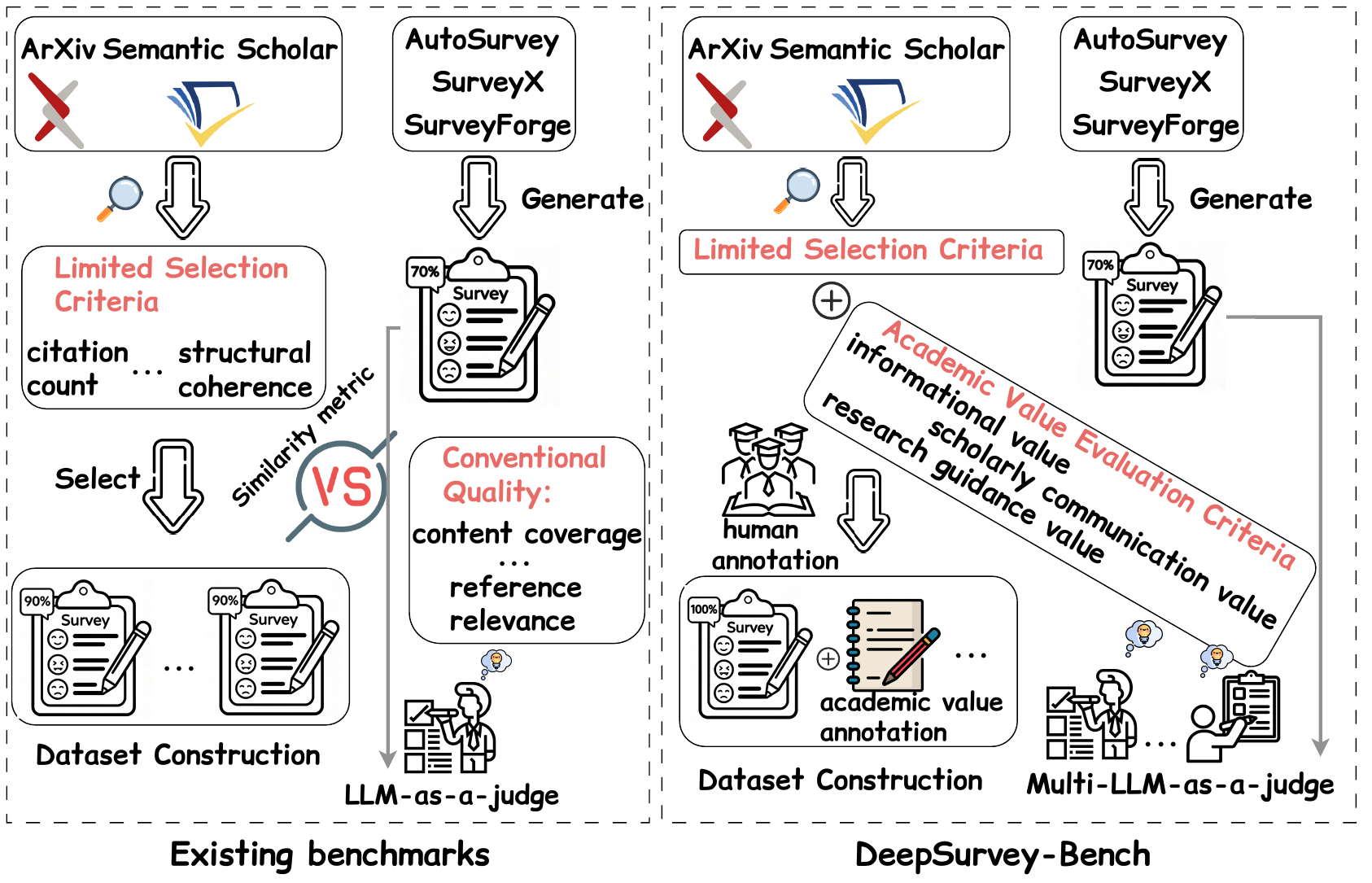}
  \caption{Comparison between existing benchmarks and our proposed DeepSurvey-Bench.}
  \label{fig:experiments}
\end{figure}

In recent years, the rapid development of artificial intelligence \cite{huynh2023artificial,wang2023scientific} and the exponential growth of research papers have made it increasingly difficult for researchers to efficiently organize vast amounts of information, resulting in greater challenges in writing surveys manually. The emergence of large language models (LLMs) \cite{achiam2023gpt,team2024qwen2,comanici2025gemini} provides a promising solution to this challenge. By integrating the powerful text generation capabilities of LLMs with retrieval-augmented generation (RAG) technology \cite{fan2024survey,chang2025main,chen2026you}, LLM-based systems can automate both literature retrieval and scientific survey generation.
Automated scientific survey generation \cite{wang2024autosurvey,wen2025interactivesurvey,chen2025surveygen,go2026lira} aims to generate comprehensive surveys for specific research topics by emulating the human survey writing process, which involves literature retrieval, outline construction, and knowledge synthesis. The complexity of this generation process highlights the urgent need for comprehensive and reliable benchmarks to evaluate the quality of generated surveys, thereby enabling a more accurate assessment of generation systems.

Existing benchmarks \cite{shi2025scisage,wang2025llm,guo2026surveylens} select human-written surveys from widely used academic resource platforms, such as arXiv\footnote{\url{https://arxiv.org/}} and Semantic Scholar\footnote{\url{https://www.semanticscholar.org/}}, based on selection criteria such as citation counts and structural coherence, to construct ground-truth survey datasets.
The automatically generated surveys are then compared with these ground-truth surveys using reference-based similarity metrics, such as ROUGE \cite{lin2004rouge} and BLEU \cite{papineni2002bleu}, and further evaluated under an LLM-as-a-judge \cite{zheng2023judging,desmond2025evalassist,huang2026think} paradigm based on conventional quality dimensions, including content coverage and reference relevance, as illustrated in Figure \ref{fig:experiments}.

However, these benchmarks suffer from two key limitations. The ground-truth datasets are insufficiently reliable because the selection criteria mainly identify highly cited or structurally coherent surveys, while lacking annotations of academic dimensions to verify their academic value. Moreover, the evaluation metrics primarily capture surface-level qualities of generated surveys and fail to assess essential academic value, such as identifying core research objectives and providing critical analysis of relationships among different studies. Consequently, existing benchmarks fail to provide an effective measure of the academic value of generated surveys.

The academic value of scientific surveys is reflected in multiple aspects. Expert evaluations of survey quality emphasize that surveys with academic value should be grounded in a comprehensive and carefully selected body of relevant literature,
clearly define research objectives or core questions \cite{torraco2005writing,denney2013write}, and provide in-depth comparison, evaluation, and interpretation of key studies in the field \cite{snyder2019literature}. Moreover, they should offer critical insights, identify research gaps and unresolved issues, and suggest promising future directions based on existing evidence \cite{xiao2019guidance,kraus2022literature}.

Motivated by the above findings, we propose \textbf{DeepSurvey-Bench}, a comprehensive benchmark for evaluating the academic value of automatically generated scientific surveys, to address these limitations of existing benchmarks. Specifically, our benchmark develops a set of academic value evaluation criteria by integrating expert evaluation principles for scientific surveys with the expertise of researchers experienced in survey writing. The criteria cover three dimensions: informational value, which reflects whether existing research is organized and synthesized into a well-structured and coherent body of knowledge; scholarly communication value, which assesses whether key studies are compared, evaluated, and critically analyzed; and research guidance value, which measures whether a survey identifies research gaps and proposes promising future directions. Then, we manually annotate human-written surveys using a unified 5-point scoring rubric aligned with these criteria and select surveys that receive high academic value scores, thereby constructing a reliable ground-truth dataset of human-written surveys with academic value annotations.
Furthermore, to enable a more specific and reproducible evaluation, we decompose the criteria into seven quantifiable academic value evaluation metrics, which are used to assess the academic value of automatically generated surveys through a multi-LLM-as-a-judge approach.

We conduct a comprehensive evaluation of existing automated survey generation methods on our proposed benchmark. Extensive experimental results demonstrate that DeepSurvey-Bench achieves strong alignment with human judgments in assessing survey academic value, uncovers academic value that cannot be captured by conventional surface-level metrics, and enables fine-grained diagnosis and iterative improvement of generated surveys.

In conclusion, our main contributions are as follows:
\begin{itemize}
    \item We propose DeepSurvey-Bench, a comprehensive benchmark for evaluating the academic value of automatically generated scientific surveys.
    \item We establish a comprehensive set of academic value evaluation criteria covering three dimensions, based on which we construct a dataset with academic value annotations and evaluate the academic value of generated surveys.
    \item We conduct extensive experiments to validate the reliability of DeepSurvey-Bench, demonstrating its strong alignment with human judgments and the effectiveness of its diagnostic feedback for improving generated surveys.
    
\end{itemize}

\section{Related Work}
\subsection{Survey Generation}
Early studies on automated scientific survey generation primarily relied on multi-document summarization techniques to generate survey sections from collections of research papers \cite{jiang2019hsds,erera2019summarization,pasunuru2021data,chen2021capturing}. In recent years, the emergence of LLMs creates new opportunities for automated survey generation. Many recent studies propose end-to-end pipelines that integrate RAG and multi-agent strategies \cite{li2024survey,liu2026llm} to enable the automated generation of long-form surveys \cite{wang2024autosurvey,liang2025surveyx,yan2025surveyforge}. AutoSurvey \cite{wang2024autosurvey} adopts a two-stage generation strategy: it first retrieves relevant literature to construct detailed outlines and then employs multiple LLMs in parallel to generate individual chapters, which are subsequently integrated into a coherent survey. 
Similarly, SurveyX \cite{liang2025surveyx} decomposes the survey generation process into preparation and generation stages, incorporating online literature retrieval, AttributeTree-based preprocessing, and an iterative polishing process to improve generation quality.
SurveyForge \cite{yan2025surveyforge} further enhances outline quality and citation accuracy by analyzing the organizational patterns of manually written outlines and integrating them with retrieved literature, while leveraging academic navigation agents to identify high-quality papers. 

\begin{figure*}[t]
  \centering
  \includegraphics[width=1.8\columnwidth]{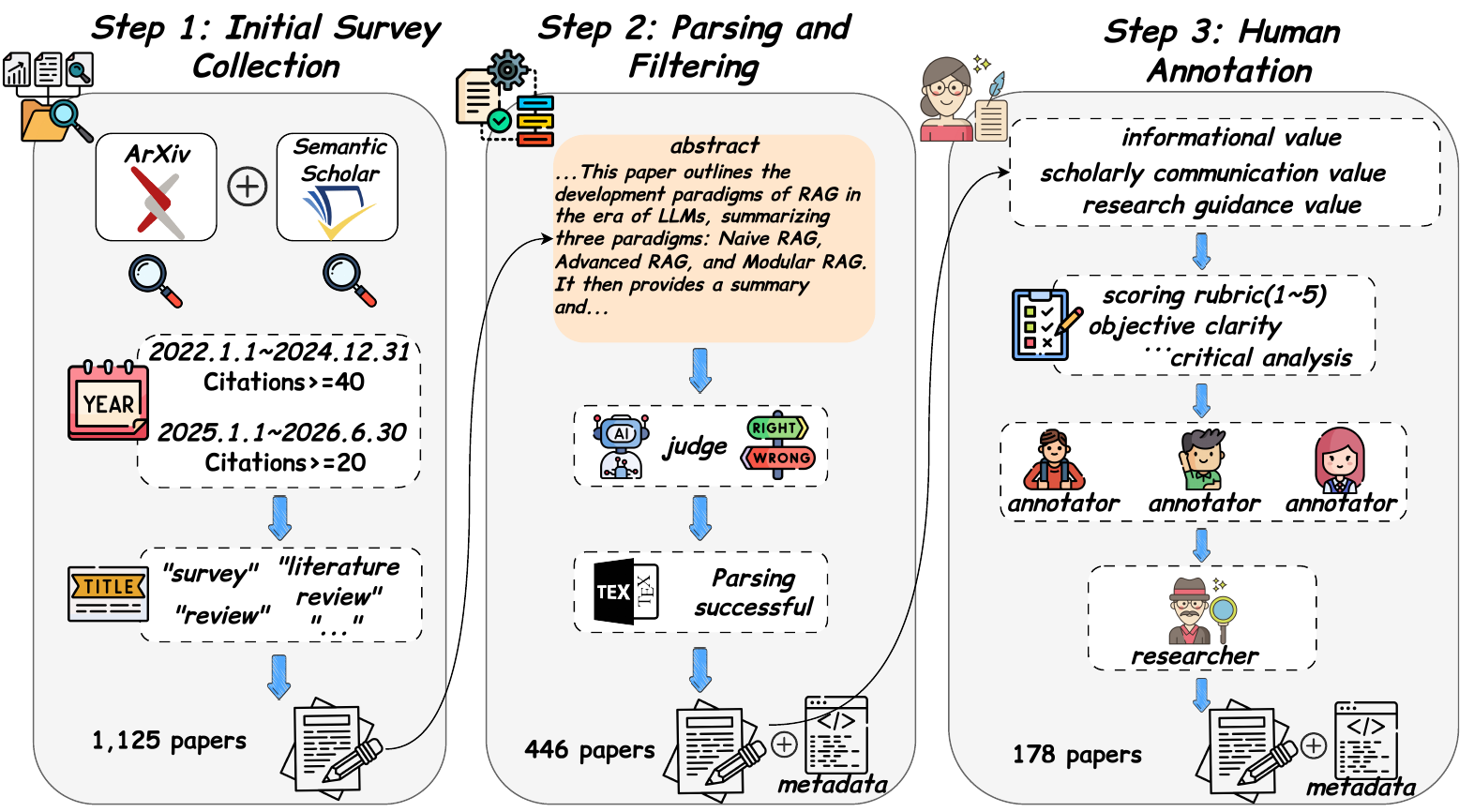}
  \caption{The dataset construction pipeline of our proposed DeepSurvey-Bench.}
  \label{fig:dataset}
\end{figure*}
\subsection{Evaluation for Survey Generation}
Existing benchmarks typically use human-written surveys as the gold standard and evaluate the quality of generated surveys based on the LLM-as-a-judge paradigm. For example,
SurveyBench \citep{yan2025surveyforge} directly selects human-written surveys based on researchers’ experience and evaluates generated surveys mainly in terms of outline structure and content relevance. 
In contrast, SurveyScope \citep{shi2025scisage} constructs its gold standard by combining objective metrics such as publication time and citation count with researchers’ judgment, and evaluates the content quality and citation accuracy of generated surveys. 
In addition, SurGE \citep{su2025surge} further incorporates human annotations based on criteria such as citation count and content coverage, and evaluates the generated surveys from multiple dimensions including citation accuracy and structural organization.

However, these benchmarks lack human annotation for the academic dimensions of the surveys, and the evaluation metrics only assess the surface-level quality and fail to evaluate the academic value of generated surveys.

\section{Survey Generation Task Definition}
The objective of automated scientific survey generation is to generate a survey $\mathcal{S}$ given a topic description $t$ and a large academic paper corpus $ R = \{ {r_1},{r_2},...,{r_n}\}$. The generated survey aims to comprehensively introduce, analyze, and synthesize research findings within a given field over a specific period. 
The generation process generally consists of the following three stages: (i) literature retrieval stage, which first retrieves an initial set of papers related to the topic $t$ and then filters them according to predefined criteria to obtain the most relevant set of reference papers ${R_s} \subseteq R$; (ii) outline generation stage, which generates a well-structured outline $O = \{ {o_1},{o_2},...,{o_m}\}$ based on the topic $t$ and the reference set ${R_s}$; (iii) survey generation stage, which generates the final survey $\mathcal{S}$ based on the topic $t$, the outline $O$, and the reference set ${R_s}$, with corresponding in-text citations and a reference list.

\section{DeepSurvey-Bench}

To address the limitations of existing benchmarks in constructing reliable academic value-annotated datasets and evaluating the academic value of generated surveys, we propose a comprehensive benchmark, named DeepSurvey-Bench. Figure \ref{fig:dataset} illustrates the dataset construction process, which consists of three steps: Initial Survey Collection (\S\ref{ini}), Parsing and Filtering (\S\ref{parse}) and Human Annotation (\S\ref{anno}).

\begin{table*}[t]
\centering
\small 
\resizebox{16cm}{!}  
{
\begin{tabular}{lccccccc}

\toprule
\textbf{Dataset} & \textbf{Domains} & \makecell[c] {\#\textbf{Survey} \\\textbf{Nums}} & \makecell[r]{\#\textbf{Average}\\\textbf{Citations}} & \makecell[c]{\#\textbf{Metadata}\\\textbf{Type Nums}} & \makecell[c]{\textbf{Academic Value}\\\textbf{Annotation}} & \makecell[c]{\textbf{Surface-Level Quality} \\\textbf{Evaluation}} & \makecell[c]{\textbf{Academic Value} \\\textbf{Evaluation}} \\ \hline
\midrule
\textbf{SurveyScope}          & CS    & 46  & 195.87   & 10          & \xmark & \cmark & \xmark \\
\textbf{SurveyBench}            & CS  &100   &  220.17  & 7       & \xmark & \cmark & \xmark \\
\textbf{SurGE}       & CS     & 205    & 275.80    & 9     & \xmark & \cmark & \xmark \\

\hline
\midrule

\rowcolor{blue!12}
\textbf{DeepSurvey-Bench} & Mixed  & 178  &235.69 & 14    & \cmark  & \cmark  & \cmark  \\
\bottomrule
\end{tabular}
}
\caption{Comparison with existing survey datasets. CS denotes Computer Science, and Mixed indicates coverage spanning both Computer Science and other disciplines.
}
\label{tab:dataset}
\end{table*}
\subsection{Initial Survey Collection}
\label{ini}
We collect scientific surveys from two widely used academic resource platforms, arXiv and Semantic Scholar. To ensure the timeliness and relevance of the selected papers, we identify surveys published between January 1, 2022, and June 30, 2026, whose titles explicitly contain the terms "survey", "literature review", "review" or "overview", thereby focusing on recent research trends and the latest developments in the field. 
Since some surveys are simultaneously indexed on both platforms, we remove duplicate entries and retain only one copy of each survey. The detailed deduplication rules are described in Appendix A.1. During the collection process, we adopt differentiated citation thresholds based on publication year, balancing the use of citation counts as a reliable indicator of academic influence and recognition with the consideration that recently published surveys may not have accumulated sufficient citations. This initial collection process is illustrated in \textit{\textbf{Step 1}} of Figure \ref{fig:dataset}. Finally, we obtain 1,125 papers preliminarily identified as candidate surveys after the initial collection stage.

\subsection{Parsing and Filtering}
\label{parse}

Since title-based filtering may still include non-survey papers, inspired by prior work \cite{bao2025surveygen}, we prompt an LLM to determine whether each paper is a genuine survey based on its abstract and predefined identification criteria, further filtering out those that contain terms such as "survey" or "review" in the title but are not genuine surveys. Then, we parse the retained surveys to extract their full texts and basic metadata, including authors, citation counts and publication year, and further enrich the metadata with details about the first-author and the publication venue. 
Only surveys with successfully extracted full-text content are retained for subsequent annotation. This parsing and filtering process is illustrated in \textit{\textbf{Step 2}} of Figure \ref{fig:dataset}.
After this parsing and filtering process, we retain 446 surveys that satisfy the requirements. Appendix A.2 details the parsing and filtering procedures.

\subsection{Human Annotation}
\label{anno}

After completing the two stages described above, we further refine the 446 candidate surveys through human annotation to construct a reliable benchmark dataset with academic value annotations. 
Prior to annotation, we first invite two researchers with extensive experience in survey writing to establish a set of academic value evaluation criteria.
These criteria are developed based on expert evaluation principles for scientific surveys from top-tier venues \citep{torraco2005writing,denney2013write,snyder2019literature,xiao2019guidance,kraus2022literature} and cover three dimensions: informational value, scholarly communication value, and research guidance value.
Furthermore, to enable a more specific and reproducible evaluation, we decompose the criteria into seven quantifiable academic value evaluation metrics and adopt a unified 5-point scoring rubric (details in \S\ref{academic}). 

Subsequently, three graduate students serve as the annotation team and independently evaluate each survey by assigning a score from 1 to 5 for each of the seven quantifiable academic value metrics. For each survey, an average academic value score is calculated based on the annotations from the three annotators, and the survey is labeled as either "select" or "discard" according to the predefined threshold\footnote{A survey is initially labeled as "select" if its average score is $\geq 4$; otherwise, it is labeled as "discard".}.
Surveys unanimously labeled as "select" by all three annotators are directly retained. Cases where two out of the three annotators assign the "select" label are further reviewed by the experienced researcher for final decisions, while the remaining surveys are discarded.  This human annotation process is illustrated in \textit{\textbf{Step 3}} of Figure \ref{fig:dataset}. 
In addition, to assess annotation reliability, we calculate Krippendorff's alpha coefficient among the three annotators and obtain a value of 0.84, indicating strong inter-annotator agreement. Ultimately, we construct a dataset comprising 178 surveys with academic value annotations. Detailed annotation prompts and scoring rubrics are provided in Appendix D.

\subsection{Statistics}
\label{statistics}
\begin{figure}[h]
\centering
\includegraphics[width=1\columnwidth]{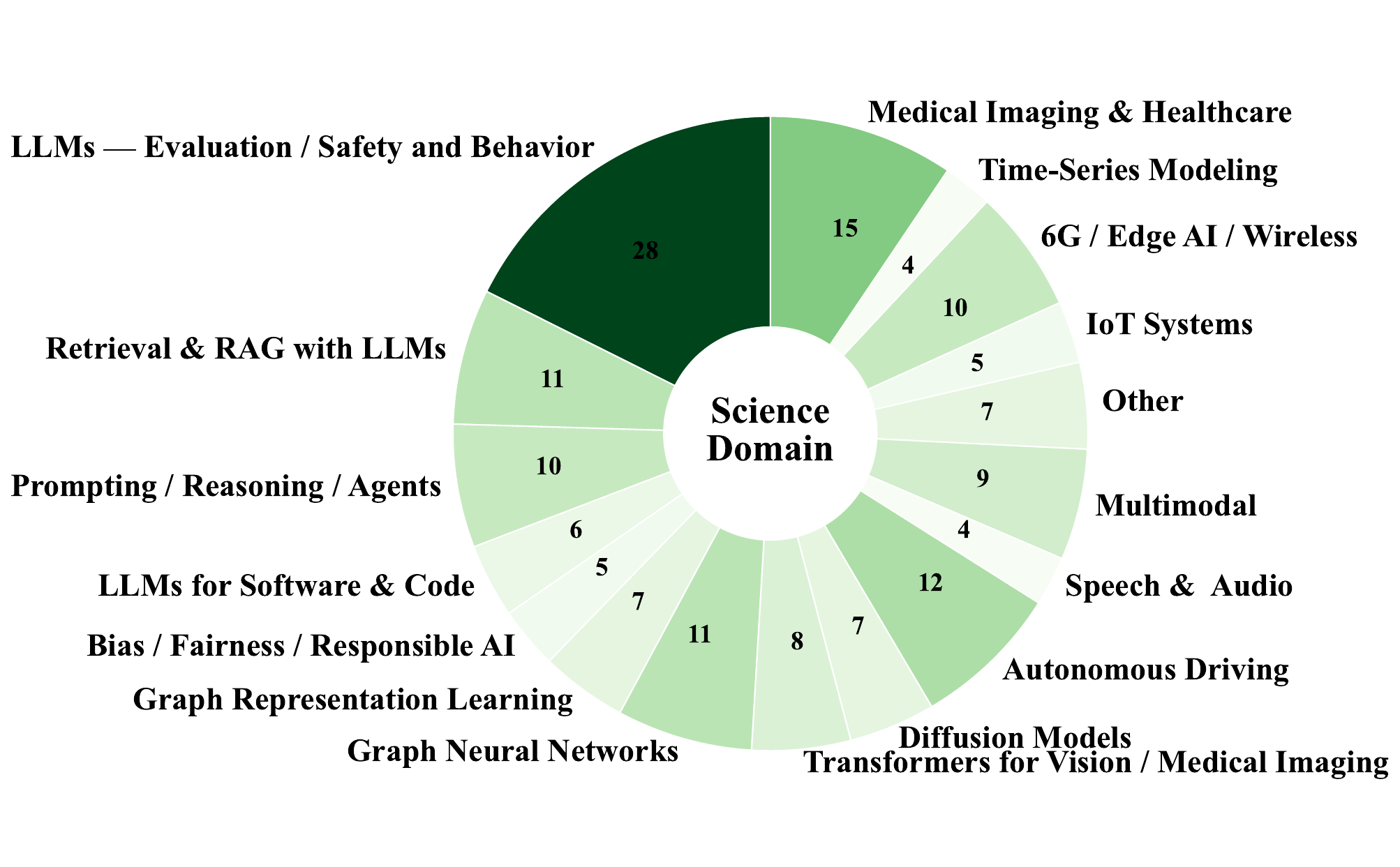} 
\caption{Distribution of DeepSurvey-Bench across different topics. Disciplines with limited survey instances are grouped into the "Other" category.}
\label{fig:sta}
\end{figure}

Our benchmark dataset contains 178 ground-truth surveys with academic value annotations, comprising 4,342 outline sections and 8,436 reference-list entries. 
The dataset format is detailed in Appendix A.3. Table \ref{tab:dataset} compares DeepSurvey-Bench with existing survey datasets. Specifically, our benchmark covers four primary disciplines: Computer Science, Physics, Medicine, and Electronic Systems, and includes 17 different topics, such as Retrieval \& RAG with LLMs and Graph Representation Learning. This broad coverage enables systematic evaluation of automated survey generation methods across diverse research domains. The distribution of different topics is shown in Figure \ref{fig:sta}.

\section{Evaluation Metrics}
\label{frame}

To comprehensively evaluate generated scientific surveys, we assess their surface-level quality while focusing primarily on their academic value. Specifically, \S\ref{sur} elaborates the evaluation metrics for surface-level quality, while \S\ref{academic} introduces the evaluation metrics for academic value.


\subsection{Surface-Level Quality}
\label{sur}
Building on the prevailing three-stage process for generating surveys \citep{wang2024autosurvey,liang2025surveyx,yan2025surveyforge}, we conduct a detailed evaluation of the surface quality of the generated surveys from three perspectives: 
\paragraph{Content Quality:}
We first employ the widely used automatic metrics \textbf{ROUGE}  \cite{lin2004rouge} and \textbf{BLEU} \cite{papineni2002bleu} in natural language generation to quantify the n-gram overlap between generated surveys and ground-truth surveys.
In addition, inspired by AutoSurvey \cite{wang2024autosurvey}, we adopt the LLM-as-a-judge approach to evaluate the content quality of generated surveys across four dimensions: \textbf{Coverage}, \textbf{Structure}, \textbf{Relevance}, and \textbf{Language Fluency}. 
The prompt and scoring rubric (1-5) are provided in Appendix D.2.
\paragraph{Outline Quality:}
A well-structured survey outline lays a solid foundation for generating a high-quality survey.
Following prior work \citep{shao2024assisting}, we introduce the quantitative metric \textbf{Heading Soft Recall (HSR)} to assess the semantic similarity between  generated and ground-truth outline headings.
In addition,
we adopt the LLM-as-a-judge paradigm to assess the outline’s \textbf{Guidance for Content Generation}, \textbf{Hierarchical Clarity}, and \textbf{Logical Coherence}. Details of the metrics are provided in Appendix B.1. The prompt and scoring rubric (1-5) are shown in Appendix D.2.
\paragraph{Reference Quality:}
References are the foundation of a scientific survey, and their selection directly determines the content composition of the survey.
For the evaluation of reference quality, we adopt the \textbf{Citation Recall} and \textbf{Citation Precision} metrics proposed by ALCE \citep{GaoYYC23} and calculate the \textbf{F1} score.
The detailed descriptions and calculation methods for these metrics are provided in Appendix B.2.
\begin{table*}[ht]
\centering
\small
\resizebox{1.95\columnwidth}{!}{%
\begin{tabular}{c|l|cccccc|cccc|ccc}
\toprule
\multirow{2}{*}{\textbf{Backbones}}
& \multirow{2}{*}{\textbf{Methods}}
& \multicolumn{6}{c|}{\textbf{Content Quality}}
& \multicolumn{4}{c|}{\textbf{Outline Quality}}
& \multicolumn{3}{c}{\textbf{Reference Quality}}
\\
\cmidrule(lr){3-8}
\cmidrule(lr){9-12}
\cmidrule(lr){13-15}

&
& \textbf{R-L}
& \textbf{B}
& \textbf{Cov.}
& \textbf{Str.}
& \textbf{Rel.}
& \textbf{LF}
& \textbf{HSR}
& \textbf{Gui.}
& \textbf{Hier.}
& \textbf{Log.}
& \textbf{CR}
& \textbf{CP}
& \textbf{F1}
\\

\midrule

\multirow{3}{*}{\textbf{GPT-4o}}
& AutoSurvey   
& 13.38 & \cellcolor{blue!35}12.13 & 4.54 & 4.28 & 4.16 & 4.14
& 63.30 & 4.00 & 4.02 & 3.98
& 52.18 & 48.47 & 50.29 \\

& SurveyX      
& \cellcolor{blue!15}13.74 & 11.34 & 4.38 & \cellcolor{blue!35}4.41 & \cellcolor{blue!35}4.36 & 4.05
& \cellcolor{blue!35}65.49 & 4.10 & 4.05 & \cellcolor{blue!35}4.20
& 53.40 & \cellcolor{blue!35}51.03 & 52.18 \\

& SurveyForge  
& \cellcolor{blue!35}14.00 & \cellcolor{blue!15}12.10 & \cellcolor{blue!35}4.67 
& \cellcolor{blue!15}4.37 & \cellcolor{blue!15}4.32 & \cellcolor{blue!15}4.21
& 64.61 & 4.08 & 4.07 & 4.07
& \cellcolor{blue!35}54.54 & \cellcolor{blue!15}50.39 & \cellcolor{blue!35}52.38\\

\midrule

\multirow{3}{*}{\textbf{Claude-3.5-haiku}}
& AutoSurvey   
& 13.47 & 11.26 & 4.05 & 4.32 & 4.15 & 4.17
& 62.98 & 4.11 & 4.08 & \cellcolor{blue!15}4.10
& \cellcolor{blue!15}54.46 & 50.14 & \cellcolor{blue!15}52.21 \\

& SurveyX      
& 13.23 & 12.04 & 4.25 & 4.24 & 4.21 & 4.00
& 64.38 & 4.02 & \cellcolor{blue!35}4.16 & 4.04
& 54.19 & 49.09 & 51.51 \\

& SurveyForge  
& 13.35 & 11.22 & 4.47 & 4.30 & 4.30 & 4.12
& 64.32 & 4.06 & 3.97 & 4.07
& 53.24 & 50.23 & 51.67 \\

\midrule
\multirow{3}{*}{\textbf{DeepSeek-v3}}
& AutoSurvey   
& 13.42 & 12.06 & 4.54 & 4.20 & 4.10 & 4.05
& 65.15 & \cellcolor{blue!35}4.18 & 4.01 & 4.03
& 51.38 & 49.24 & 50.29 \\

& SurveyX      
& 13.25 &11.13 & 4.42 & 4.26 & 4.26 & \cellcolor{blue!35}4.24
& \cellcolor{blue!15}65.30 & 4.10 & \cellcolor{blue!15}4.14 & \cellcolor{blue!15}4.10
& 52.08 & 49.31 & 50.65 \\

& SurveyForge  
& 13.68 & 11.51 & \cellcolor{blue!15}4.60 & 4.15 
& 4.20 & 4.13
& 64.63 & \cellcolor{blue!15}4.16 & 4.09 & 4.04
& 52.10 & 50.10 & 51.08 \\

\midrule
\multicolumn{2}{c|}{\textbf{Human-written Upper Bound}}
& - & - & 4.85 & 4.80 & 4.83 & 4.75
& - & 4.88 & 4.78 & 4.76
& - & - & - \\

\bottomrule
\end{tabular}}
\caption{Surface-level quality evaluation results of the generated surveys. Human-written surveys are included as an empirical upper bound reference. Metric names are abbreviated for conciseness. \colorbox{blue!35}{Best} and
 \colorbox{blue!15}{second-best} scores are highlighted. }
\label{tab:surface_quality}
\end{table*}

\renewcommand{\arraystretch}{0.7}
\begin{table*}[ht]
\centering
\small
\resizebox{1.8\columnwidth}{!}{%
\begin{tabular}{c|l|
cccc|ccc|ccc}
\toprule
\multirow{2}{*}{\textbf{Backbones}}
& \multirow{2}{*}{\textbf{Methods}}

& \multicolumn{4}{c|}{\textbf{Information}}
& \multicolumn{3}{c|}{\textbf{Scholarly com.}}
& \multicolumn{3}{c}{\textbf{Research gui.}}
\\

\cmidrule(lr){3-6} \cmidrule(lr){7-9} \cmidrule(lr){10-12}

&
& \textbf{OC}
& \textbf{CEC}
& \textbf{DMC}
& \textbf{Avg}
& \textbf{IC}
& \textbf{CA}
& \textbf{Avg}
& \textbf{RG}
& \textbf{FW}
& \textbf{Avg}
\\

\midrule

\multirow{3}{*}{\textbf{GPT-4o}}
& AutoSurvey   
& 3.62 & 3.85 & \cellcolor{blue!15}3.48 & \cellcolor{blue!15}3.65
& 3.25 & 3.37 & 3.31
& 3.62 & \cellcolor{blue!15}4.04 & 3.83 \\

& SurveyX      
& \cellcolor{blue!35}3.84 & \cellcolor{blue!35}3.98 & \cellcolor{blue!35}3.82 & \cellcolor{blue!35}3.88
& \cellcolor{blue!35}3.62 & 3.58 & \cellcolor{blue!35}3.60
& \cellcolor{blue!15}3.85 & \cellcolor{blue!35}4.12 & \cellcolor{blue!35}3.99 \\

& SurveyForge  
& 3.56 & \cellcolor{blue!15}3.90 & 3.31 & 3.59
& 3.40 & \cellcolor{blue!35}3.63 & 3.51
& 3.80 & 4.00 & 3.90 \\

\midrule

\multirow{3}{*}{\textbf{Claude-3.5-haiku}}
& AutoSurvey   
& 3.06 & 3.51 & 3.29 & 3.29
& 3.18 & 3.15 & 3.17
& 3.62 & 3.74 & 3.68 \\

& SurveyX      
& 3.62 & 3.58 & 3.27 & 3.49
& 3.26 & 3.42 & 3.34
& 3.78 & 4.01 & 3.90 \\

& SurveyForge  
& 3.41 & 3.64 & 3.10 & 3.38
& 3.23 & 3.56 & 3.40
& 3.64 & 3.86 & 3.75 \\

\midrule

\multirow{3}{*}{\textbf{DeepSeek-v3}}
& AutoSurvey   
& 3.50 & 3.67 & 3.06 & 3.41
& 3.38 & 3.38 & 3.38
& \cellcolor{blue!15}3.85 & 3.83 & 3.84 \\

& SurveyX      
& \cellcolor{blue!35}3.84 & 3.74 & 3.15 & 3.58
& \cellcolor{blue!15}3.42 & \cellcolor{blue!15}3.60 & \cellcolor{blue!15}3.51
& \cellcolor{blue!35}3.94 & 3.90 & \cellcolor{blue!15}3.92 \\

& SurveyForge  
& \cellcolor{blue!15}3.63 & 3.60 & 3.18 & 3.47
& 3.20 & 3.47 & 3.33
& 3.80 & 4.00 & 3.90 \\

\midrule
\multicolumn{2}{c|}{\textbf{Human-written Upper Bound}}
& 4.58 & 4.74 & 4.70 & 4.67
& 4.56 & 4.70 & 4.63
& 4.82 & 4.60 & 4.71 \\
\bottomrule
\end{tabular}}
\caption{Academic value evaluation results of the generated surveys. Human-written surveys are included as an empirical upper bound reference. 
Metric names are abbreviated for conciseness. \colorbox{blue!35}{Best} and \colorbox{blue!15}{second-best} scores are highlighted.}
\label{tab:academic}
\end{table*}

\subsection{Academic Value}

\label{academic}
To make the evaluation of the survey's academic value more specific and feasible, we further decompose the three dimensions (\S \ref{anno}) into seven quantifiable evaluation metrics. The prompts and scoring rubric (1-5) for these seven academic metrics are all in Appendix D.2. 
\paragraph{Informational Value:}
We further decompose informational value into three key metrics: \textbf{Objective Clarity}, \textbf{Classification-Evolution Coherence}, and \textbf{Dataset \& Metric Coverage}. 
Objective Clarity evaluates the clarity of research objectives in the survey. 
Classification–Evolution Coherence assesses the clarity of method classification and the evolution of technological development in the survey. 
Dataset \& Metric Coverage evaluates the coverage of datasets and the rationality of the evaluation metrics in the survey. 

\paragraph{Scholarly Communication Value:}
We further decompose scholarly communication value into two key metrics: 
\textbf{In-depth Comparison} and \textbf{Critical Analysis}. In-depth Comparison assesses whether the survey systematically compares multiple methods, clearly describing the advantages and disadvantages of each method.
Critical Analysis evaluates whether the survey provides critical analysis of different methods, focusing on explaining the underlying reasons for their differences.

\paragraph{Research Guidance Value:}
We further decompose research guidance value into two key metrics: \textbf{Research Gap} and \textbf{Future Work}. Research Gap evaluates whether the survey identifies and analyzes key unresolved issues and limitations in the current research field. 
Future Work evaluates whether the survey propose forward-looking future research directions based on identified research gaps and practical needs.

\label{metric}

\section{Experiments and Analysis}
\label{experiment}

\subsection{Baselines}
\label{base}

We employ three automated survey generation methods as baselines to compare their performance under our proposed evaluation metrics.
Detailed descriptions of these baselines are provided in Appendix C.1.
\subsection{Implementation Details}
We select Claude-3.5-Haiku, GPT-4o, and DeepSeek-v3 as backbone models for different survey generation methods, representing different capability levels. This allows us to examine the discriminative ability of the proposed evaluation criteria across models with varying capabilities.
For a fair comparison, each method follows the retrieval strategies and generation configurations specified in its original paper. To reduce the impact of generation variability, each survey generation experiment is repeated multiple times, and we report the average performance.
For automated evaluation, relying on a single LLM as a judge to assess generated surveys may introduce inherent biases, thereby affecting the reliability of the results. To mitigate this limitation, we adopt a multi-LLM-as-a-judge ensemble strategy \cite{li2024llms,li2025generation}, including GPT-5.2 (Open AI), Gemini-3-Pro-Preview (Google), and Claude-4-Sonnet (Anthropic). These models jointly assess each generated survey, and their average score is used as the final evaluation result\footnote{The agreement among GPT-5.2, Gemini-3-Pro-Preview, and Claude-4-Sonnet is verified in \S\ref{corre}}. 
Furthermore, to ensure reproducible evaluation, we set the temperature parameter to 0 for all LLM evaluators during the evaluation process.
\label{imple}

\subsection{Experimental Results}
\label{experiment results}
We evaluate three survey generation baselines with different backbone models and report the evaluation results in terms of both surface-level quality and academic value. Table \ref{tab:surface_quality} reports the surface-level quality evaluation results, while Table \ref{tab:academic} presents the academic value evaluation results\footnote{We include human-written surveys as a reference upper bound in the last row of the table to indicate the performance gap between generated and human-written surveys.}. Based on these results, we make the following observations:

First, regardless of whether evaluated by surface quality or academic value metrics, surveys generated with GPT-4o as the backbone generally achieve better performance than those generated with other backbones. Specifically, GPT-4o-based methods achieve the best results on 10 out of 13 surface quality metrics and 6 out of 7 academic value metrics, demonstrating their consistent advantages in both conventional survey quality and deeper academic value assessment. This indicates that stronger backbone models can leverage their powerful language understanding and reasoning capabilities to generate high-quality surveys with academic value.

Second, different generation methods exhibit distinct advantages on surface-level quality metrics. For example, SurveyX achieves the highest scores in Structure (4.41) and Relevance (4.36), SurveyForge performs best in ROUGE-L (14.00) and Coverage (4.67), while AutoSurvey achieves the highest BLEU score (12.13) and Guidance for Content Generation score (4.18). In contrast, our academic value evaluation metrics reveal more distinctive differences among different methods. For example, SurveyX achieves the best performance on all academic value metrics except CA (Critical Analysis), indicating that our academic value evaluation criteria can clearly and robustly distinguish the capabilities of different methods in generating surveys with academic value.

Third, although most generated surveys achieve relatively high scores on LLM-based surface quality metrics, with most content and outline quality scores exceeding 4.0, their academic value scores generally remain below 4.0 across the three dimensions. This discrepancy indicates that existing survey generation methods can effectively imitate the surface characteristics of human-written surveys, but still struggle to capture deeper academic value.  It also highlights the necessity of evaluating academic value beyond conventional quality metrics, which is further validated by the correlation analysis presented in \S\ref{Correlation}. 
In addition, the generation time for baselines and API costs are shown in the Appendix C.2.

\subsection{Human Evaluation and Inter-Evaluator Agreement Analysis}
\label{corre}

Although the multi-LLM-as-a-judge paradigm provides an efficient approach for evaluating generated surveys, relying solely on LLM-based evaluation may raise concerns regarding evaluator reliability and potential model preferences toward AI-generated content. To alleviate this concern, we randomly sample 40 generated surveys covering diverse research topics and invite three evaluators (two doctoral students and one experienced researcher, all with experience in survey writing) to assess these surveys using the same academic value scoring rubric as the LLM evaluators. The detailed human evaluation results are provided in Appendix C.3. Subsequently, based on both LLM-based and human evaluation results, we conduct a comprehensive inter-evaluator agreement analysis from three complementary perspectives: LLM–LLM agreement, human–human agreement, and human–LLM agreement. Specifically, for each of the seven academic value evaluation metrics defined in \S\ref{academic}, we calculate Cohen’s Kappa \cite{thakur2025judging} between different evaluator pairs. The corresponding agreement results are summarized in Table \ref{tab:all}. The average agreement scores for LLM–LLM, human–human, and human–LLM comparisons all exceed 0.80 across the seven academic value evaluation metrics. These consistent agreement results across different evaluator groups demonstrate the reliability of the evaluation process and validate the stability and reproducibility of the proposed evaluation criteria.

\begin{table}[t]
\caption{Consistency statistics across LLM evaluators, human evaluators, and human--LLM evaluators.}
\label{tab:all}
\centering
\resizebox{\columnwidth}{!}{
\begin{tabular}{lcccccccc}
\toprule
\multirow{2}{*}{\bf Evaluators} & \multicolumn{8}{c}{\bf Academic Value } \\
\cmidrule(lr){2-9}
& OC & CEC & DMC & IC & CA & RG & FW & Avg \\
\midrule

\multicolumn{9}{c}{\bf LLM vs. LLM} \\
\midrule
GPT vs. Gemini & 0.80 & 0.78 & 0.84 & 0.77 & 0.82 & 0.83 & 0.80 & 0.81 \\
GPT vs. Claude & 0.86 & 0.85 & 0.80 & 0.84 & 0.83 & 0.78 & 0.79 & 0.82 \\
Gemini vs. Claude & 0.80 & 0.81 & 0.78 & 0.80 & 0.82 & 0.82 & 0.84 & 0.81 \\

\midrule
\multicolumn{9}{c}{\bf Human vs. Human} \\
\midrule
Evaluator 1 vs. Evaluator 2 & 0.80 & 0.85 & 0.82 & 0.86 & 0.82 & 0.80 & 0.78 & 0.82 \\
Evaluator 1 vs. Evaluator 3 & 0.83 & 0.84 & 0.88 & 0.83 & 0.78 & 0.84 & 0.80 & 0.83 \\
Evaluator 2 vs. Evaluator 3 & 0.84 & 0.84 & 0.85 & 0.88 & 0.80 & 0.89 & 0.84 & 0.85 \\

\midrule
\multicolumn{9}{c}{\bf Human vs. LLM} \\
\midrule
GPT-5.2 & 0.91 & 0.83 & 0.85 & 0.85 & 0.88 & 0.81 & 0.85 & 0.85 \\
Gemini-3-Pro-Preview & 0.83 & 0.86 & 0.90 & 0.84 & 0.87 & 0.85 & 0.84 &  0.86\\
Claude-4-Sonnet & 0.83 & 0.82 & 0.89 & 0.76 & 0.87 & 0.90 & 0.88 & 0.85 \\

\bottomrule
\end{tabular}
}
\end{table}

\begin{table}[t]
\centering
\small
\begin{tabular}{p{0.3\columnwidth}lcc}
\toprule
\textbf{Benchmarks}
& \textbf{Kendall $\tau$}
& \textbf{Spearman $\rho$}
\\
\midrule

SurveyScope
& 0.36
& 0.44
\\

SurveyBench 
& 0.45
& 0.48
\\

SurGE
& 0.41
& 0.40
\\

\bottomrule
\end{tabular}
\caption{
Ranking correlation among different benchmarks.
}
\label{tab:rank}
\end{table}

\subsection{Correlation Analysis among Survey Evaluation Benchmarks}
\label{Correlation}
To investigate whether the evaluation results of existing survey benchmarks are consistent with the academic value assessment of 
DeepSurvey-Bench, we conduct a correlation analysis among different evaluation benchmarks. Specifically, we randomly sample 40 generated surveys from different backbone models and generation baselines, and compare the quality ranking results produced by 
existing benchmarks and DeepSurvey-Bench on these surveys. We then calculate Kendall and Spearman coefficients to measure the ranking correlation between different benchmarks. As shown in Table \ref{tab:rank}, existing benchmarks exhibit moderate ranking correlations ($0.36 \leq \tau \leq 0.45$, $0.40 \leq \rho \leq 0.48$) with DeepSurvey-Bench. 
This indicates that existing benchmarks capture only part of the quality aspects considered by DeepSurvey-Bench, which further evaluates the academic value of generated surveys.
Furthermore, we conduct a metric-level correlation analysis between surface-level quality and overall academic value. Specifically, we use the surface-level quality and academic value scores of the above 40 samples reported in \S\ref{experiment results}, and compute Pearson and Spearman correlation coefficients between each surface-level quality dimension and the overall academic value score.
\begin{table}[t]
\centering
\small
\begin{tabular}{lcc}
\toprule
\textbf{Surface-Level Quality}
& \textbf{Pearson $r$}
& \textbf{Spearman $\rho$}
\\
\midrule

Outline Quality
& 0.24
& 0.22
\\

Content Quality
& 0.36
& 0.35
\\

Reference Quality
& 0.20
& 0.18
\\

\bottomrule
\end{tabular}
\caption{
Correlation between surface-level quality and academic value.
}
\label{tab:cor}
\end{table} 
As shown in Table \ref{tab:cor}, all surface-level quality perspectives exhibit weak correlations with academic value, with both coefficients remaining below 0.40. These results indicate that surface-level quality metrics provide limited insights into the academic value of generated surveys and fail to capture their academic contributions. 

\begin{table}[t]
\centering
\small
\begin{tabular}{p{0.60\columnwidth}c}
\toprule
\textbf{Evidence Integrity Metric}
& \textbf{Rate}\\
\midrule
Factual Correctness & 94\%\\
Claim-Evidence Alignment & 91\%\\
Citation Reliability & 93\%\\
\bottomrule
\end{tabular}

\caption{
Evidence integrity validation results of 40 generated surveys with high academic value scores.
}
\label{tab:evidence}
\end{table}

\subsection{Evidence Validation and Benchmark-Guided Improvement}
To further validate the reliability and practical utility of DeepSurvey-Bench, we conduct two analyses: evidence integrity validation and benchmark-guided optimization. 
First, we conduct an evidence integrity validation to verify whether the academic value evaluation reflects substantive academic quality rather than merely stylistic or rhetorical characteristics. Specifically, we randomly select 40 generated surveys with high academic value scores and manually examine them from three perspectives: factual correctness, claim-evidence alignment, and citation reliability. 
As shown in Table \ref{tab:evidence}, the selected surveys achieve high evidence integrity (91\%--94\%), demonstrating that surveys with high academic value scores generally exhibit strong evidence integrity. 
Furthermore, we evaluate the utility of DeepSurvey-Bench for survey refinement. Specifically, we select 40 surveys with low academic value scores and refine them using the diagnostic feedback provided by academic value evaluation.
\begin{figure}[ht]
\centering
\includegraphics[width=0.9\columnwidth]{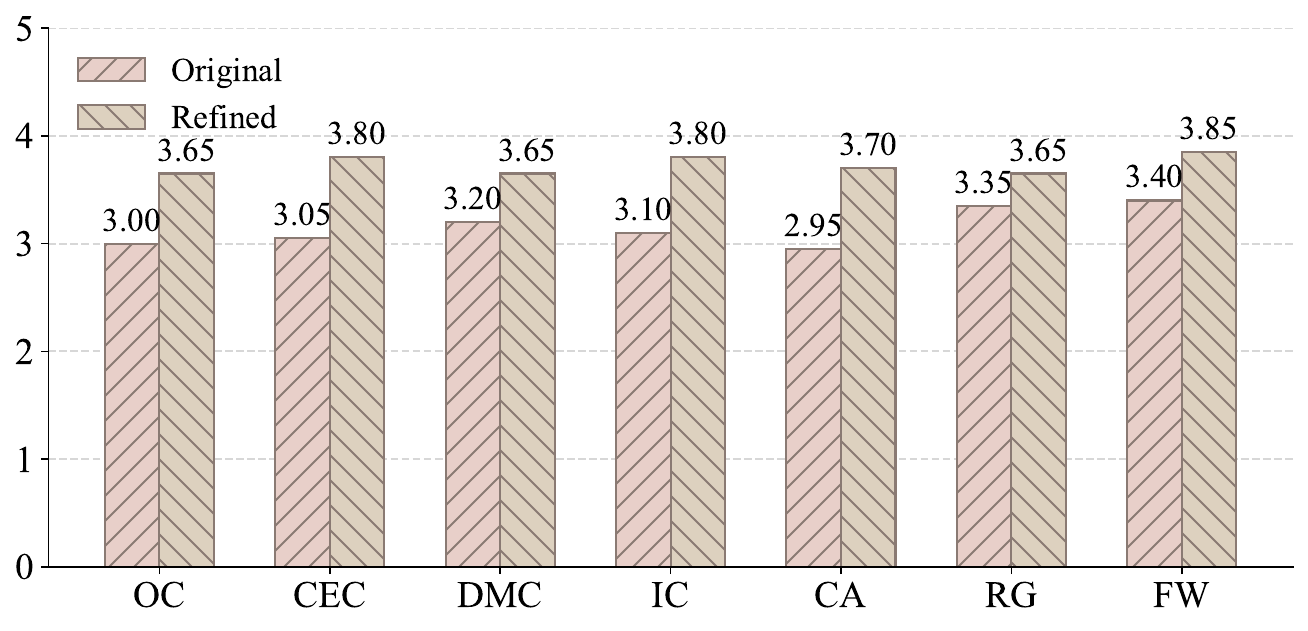} 
\caption{Comparison of academic value evaluation results between original and benchmark-guided refined surveys.}
\label{fig:optimization}
\end{figure}
As shown in Figure \ref{fig:optimization}, the refined surveys achieve improvements across all academic value metrics under the LLM ensemble evaluation compared with the original surveys. These results indicate that DeepSurvey-Bench can provide effective guidance for improving generated surveys and support the optimization of survey generation approaches. Detailed validation and refinement experiments are provided in Appendix C.4.

\section{Conclusion and Future Work}
In this paper, we propose DeepSurvey-Bench, a comprehensive benchmark for evaluating the academic value of automatically generated scientific surveys. DeepSurvey-Bench establishes comprehensive academic value evaluation criteria, constructs a reliable dataset with academic value annotations based on these criteria, and evaluates generated surveys across three academic dimensions.
Extensive experimental results demonstrate that DeepSurvey-Bench achieves strong alignment with human judgments in evaluating the academic value of surveys, offers reliable assessments of academic value, and enables fine-grained diagnosis and iterative improvement of generated surveys.
In future work, we plan to improve inter-paper relation modeling and literature analysis to enhance the academic value of generated surveys.

\bibliography{aaai2027}

\clearpage
\section{Appendix}
\paragraph{A \quad DeepSurvey-Bench}

\paragraph{A.1 \quad Initial Survey Collection.}

Since arXiv and Semantic Scholar have overlapping coverage, the same survey may be indexed by both platforms or appear as multiple versions. We therefore deduplicate the collected records using persistent identifiers, primarily DOI and arXiv ID.
The initial retrieval of the two platforms yields 1,367 records. We first compare the DOI information associated with each record. Records sharing the same DOI are considered different index entries from the same paper and are merged into a single record. This step removes 176 duplicate records, leaving 1,191 records.

We then identify duplicates using arXiv IDs. To account for different revisions of the same preprint, version suffixes are removed before matching. For example, \texttt{arXiv:xxxx.xxxxxv1} and \texttt{arXiv:xxxx.xxxxxv2} are both mapped to the same base identifier, \texttt{arXiv:xxxx.xxxxx}. Records sharing the same base arXiv ID are therefore regarded as the same work. This step removes a further 66 duplicate records. When a preprint and a formally published version share the same DOI or base arXiv ID, we retain the formally published version and remove the corresponding preprint record. The published version is preferred because it generally represents the final version of the paper and contains more complete and standardized bibliographic information. If no formally published version is available, the latest arXiv version is retained.

In total, 242 duplicate records are removed from the initial 1,367 records. After deduplication, 1,125 unique papers are retained and preliminarily identified as candidate surveys for the subsequent dataset construction process.

\paragraph{A.2 \quad Parsing and Filtering.}

Specifically, GPT-4o is provided with the title and abstract of each candidate paper and instructed to determine whether the paper primarily provides a structured synthesis of multiple existing studies. Papers are retained if they systematically organize and analyze prior research through literature synthesis, taxonomy construction, methodological comparison, or discussions of research challenges and future directions. By contrast, papers primarily reporting questionnaire-based studies or introducing new methods, systems, datasets, or benchmarks are excluded, even if their titles contain terms such as survey,"review", or "overview". This procedure removes 598 false-positive records, leaving 527 scientific surveys.

For each retained survey, we retrieve the formally published full text when available; otherwise, the latest arXiv version is used. The retrieved PDF or HTML files are converted into structured text by extracting the title, abstract, section hierarchy, paragraph-level content, in-text citations, and reference list. During preprocessing, repeated headers, footers, page numbers, and other layout artifacts are removed, while character encoding, whitespace, line breaks, and cross-line hyphenation are normalized. Section hierarchies and paragraph boundaries are preserved to support subsequent academic-value annotation. Among the 527 surveys, 53 are excluded because their full texts are unavailable, inaccessible, or cannot be obtained in a usable format. The remaining 474 documents are further subjected to parsing quality verification. A document is excluded if its main body is missing or severely truncated, substantial portions of the extracted content are unreadable, its section structure cannot be reliably recovered, or the parsed text consists predominantly of references, tables, or layout artifacts. For each retained survey, we also collect and standardize bibliographic metadata, including the author list, publication year, citation count, DOI, arXiv ID, first-author information, and publication venue. These metadata are not used as additional filtering criteria but support dataset characterization and distribution analysis, including analyses of temporal coverage, citation impact, and publication sources. After completing the above procedures, 446 surveys are retained for subsequent academic value annotation\footnote{To facilitate reproducibility, all code and data have been submitted as part of the Code and Data Supplement and will be publicly released upon acceptance.}.

\paragraph{A.3 \quad Statistics.}
The format of the dataset is shown in Figure \ref{fig: dataformat}, and each field is briefly described in Table \ref{tab:metada}.

\begin{figure}[ht] 
\vspace{-1em}
\begin{AIbox}{DeepSurvey-Bench Data Format}
{\color{black}} 
\begin{lstlisting}[style=prompt]
{
  "authors": [
    "string"
  ],
  "literature_review_title": "string",
  "year": "string",
  "date": "string",
  "category": "string",
  "abstract": "string",
  "structure": [
    {
      "section_title": "string",
      "level": "string",
      "content": "string",
      "origin_cites_number": "number"
    }
  ],
  "literature_review_id": "number",
  "meta_info": {
    "cite_counts": "number",
    "venue": "string",
    "influential_citation_count": "number",
    "Author_info": {
      "Publicationsh": "number",
      "h_index": "number",
      "Citations": "number",
      "Highly Influential Citations": "number"
    },
    "all_cites_title": [
      "string"
    ]
  }
}
\end{lstlisting}
\end{AIbox} 
\caption{The data format of DeepSurvey-Bench.}
\label{fig: dataformat}
\vspace{-1em}
\end{figure}

\begin{table}[ht]
\centering
\small
\resizebox{\columnwidth}{!}{%
\begin{tabular}{ll} 
\toprule
\textbf{Metadata}              & \textbf{Description}  \\ 
\midrule

authors              &  List of contributing researchers. \\
literature\_review\_title              & The title of the survey.      \\  
year               & The publication year of the survey.     \\   
date       &          The timestamp of publication.             \\      
category       &  The classification or type of the survey.     \\
abstract       & A brief summary of the survey's content, aims, and findings.  \\
literature\_review\_id    & A unique identifier for the survey. \\
cite\_counts                   &  The total number of citations for the survey.           \\  
conference\_journal\_name     & The conference or journal where the survey was published.        \\   
influential\_citation\_count  & The number of influential citations received by the survey. \\
publication                    &    The total number of papers published by the first author.   \\
h\_index                     & The h-index of the first author. \\
citations                    &   The total number of citations for the first author’s work.    \\
all\_cites\_title   &  The titles of all the references cited in the survey.  \\
\bottomrule
\end{tabular}
}
\caption{Metadata fields and descriptions in the DeepSurvey-Bench dataset.}
\label{tab:metada}
\end{table}

\paragraph{B \quad Evaluation Metrics}

\paragraph{B.1 \quad Outline Quality.}

To calculate Heading Soft Recall (HSR), we flatten the multi-level outline of each survey into a set of heading texts. Let $P$ denote the headings in the generated outline and $G$ denote those in the corresponding human-written survey. HSR measures the proportion of the semantic information contained in the ground-truth headings that is covered by the generated headings.

Given a heading set $A=\{A_i\}_{i=1}^{K}$, the soft count of each heading $A_i$ is defined as
\begin{equation}
\operatorname{count}(A_i)
=
\frac{1}
{\sum_{j=1}^{K}\operatorname{Sim}(A_i,A_j)},
\label{eq:soft_count}
\end{equation}
where the semantic similarity between two headings is computed as
\begin{equation}
\operatorname{Sim}(A_i,A_j)
=
\cos\left(
\operatorname{embed}(A_i),
\operatorname{embed}(A_j)
\right).
\label{eq:heading_similarity}
\end{equation}
Here, $\operatorname{embed}(\cdot)$ denotes the sentence representation obtained using \texttt{paraphrase-MiniLM-L6-v2}. The soft count assigns a lower contribution to headings that are semantically redundant with other headings in the same set, thereby preventing multiple similar headings from being repeatedly counted.

The soft cardinality of a heading set is then calculated as
\begin{equation}
\operatorname{card}(A)
=
\sum_{i=1}^{K}\operatorname{count}(A_i).
\label{eq:soft_cardinality}
\end{equation}
Based on the soft cardinality, HSR is defined as
\begin{equation}
\operatorname{HSR}(G,P)
=
\frac{\operatorname{card}(G\cap P)}
{\operatorname{card}(G)},
\label{eq:hsr}
\end{equation}
where the soft cardinality of the intersection is derived using the inclusion--exclusion principle:
\begin{equation}
\operatorname{card}(G\cap P)
=
\operatorname{card}(G)
+
\operatorname{card}(P)
-
\operatorname{card}(G\cup P).
\label{eq:soft_intersection}
\end{equation}
The union $G\cup P$ contains all headings from the generated and ground-truth outlines, and its soft cardinality is computed using Equations~\ref{eq:soft_count}--\ref{eq:soft_cardinality}. A higher HSR indicates that the generated outline provides greater semantic coverage of the headings in the human-written survey, while accounting for paraphrasing and semantic redundancy.

\renewcommand{\arraystretch}{0.7}
\begin{table*}[ht]
\centering
\small
\resizebox{1.8\columnwidth}{!}{%
\begin{tabular}{c|l|
cccc|ccc|ccc}
\toprule
\multirow{2}{*}{\textbf{Backbones}}
& \multirow{2}{*}{\textbf{Methods}}

& \multicolumn{4}{c|}{\textbf{Information}}
& \multicolumn{3}{c|}{\textbf{Scholarly com.}}
& \multicolumn{3}{c}{\textbf{Research gui.}}
\\

\cmidrule(lr){3-6} \cmidrule(lr){7-9} \cmidrule(lr){10-12}

&
& \textbf{OC}
& \textbf{CEC}
& \textbf{DMC}
& \textbf{Avg}
& \textbf{IC}
& \textbf{CA}
& \textbf{Avg}
& \textbf{RG}
& \textbf{FW}
& \textbf{Avg}
\\

\midrule

\multirow{3}{*}{\textbf{GPT-4o}}
& AutoSurvey   
& 3.50 & 3.72 & 3.40 & 3.54
& 3.24 & 3.30 & 3.27
& 3.54 & 3.92 & 3.73 \\

& SurveyX      
& \cellcolor{blue!35}3.78 & \cellcolor{blue!35}3.88 & \cellcolor{blue!35}3.76 &\cellcolor{blue!35}3.81
& \cellcolor{blue!35}3.56 & 3.47 & \cellcolor{blue!35}3.52
& \cellcolor{blue!35}3.80 & \cellcolor{blue!35}4.00 & \cellcolor{blue!35}3.90 \\

& SurveyForge  
& 3.50 & 3.82 & 3.28 & 3.53
& 3.38 & \cellcolor{blue!35}3.52 & 3.45
& 3.69 & 3.92 & 3.81 \\
\bottomrule
\end{tabular}}
\caption{Human evaluation results for GPT-4o based survey generation methods. Metric names are abbreviated for conciseness. \colorbox{blue!35}{Best} score is highlighted.}
\label{tab:human}
\end{table*}

\paragraph{B.2 \quad Reference Quality.}
Following ALCE \citep{GaoYYC23}, we evaluate reference quality using Citation Recall (CR), Citation Precision (CP), and their harmonic mean, F1. Specifically, we decompose each generated survey into a set of citation-worthy claims
$C=\{c_i\}_{i=1}^{|C|}$. For each claim $c_i$, let
$R_i=\{r_{ik}\}_{k=1}^{|R_i|}$ denote the set of cited evidence associated with it, where $r_{ik}$ represents the evidence provided by the $k$-th cited reference. We employ a natural language inference model $h$ to determine whether a claim is supported by a given set of cited evidence:

\begin{equation}
h(c_i,R_i)=
\begin{cases}
1, & \text{if } R_i \text{ supports } c_i,\\
0, & \text{otherwise}.
\end{cases}
\end{equation}

Citation Recall measures the proportion of claims that are supported by their associated citations:

\begin{equation}
\mathrm{CR}
=
\frac{1}{|C|}
\sum_{i=1}^{|C|}
h(c_i,R_i).
\label{eq:citation_recall}
\end{equation}

A higher Citation Recall indicates that a larger proportion of the claims in the generated survey are supported by the cited evidence. For a claim without any associated citation, we set $R_i=\varnothing$ and $h(c_i,R_i)=0$.

Citation Precision evaluates whether each individual citation is relevant or necessary for supporting its associated claim. Following \citet{GaoYYC23}, we define

\begin{equation}
g(c_i,r_{ik})
=
\mathbb{I}
\left[
h(c_i,\{r_{ik}\})=1
\;\lor\;
h(c_i,R_i\setminus\{r_{ik}\})=0
\right],
\label{eq:citation_relevance}
\end{equation}

where $\mathbb{I}[\cdot]$ is the indicator function. A citation is considered valid if it independently supports the claim or if removing it causes the remaining citations to no longer support the claim. Citation Precision is then computed as

\begin{equation}
\mathrm{CP}
=
\frac{
\displaystyle
\sum_{i=1}^{|C|}
\sum_{k=1}^{|R_i|}
h(c_i,R_i)\,
g(c_i,r_{ik})
}{
\displaystyle
\sum_{i=1}^{|C|}
|R_i|
}.
\label{eq:citation_precision}
\end{equation}

The term $h(c_i,R_i)$ ensures that individual citations receive credit only when the complete citation set supports the corresponding claim. A higher Citation Precision indicates that fewer irrelevant or redundant citations are included.

Finally, the F1 score summarizes Citation Recall and Citation Precision through their harmonic mean:

\begin{equation}
\mathrm{F1}
=
\frac{2\cdot \mathrm{CR}\cdot \mathrm{CP}}
{\mathrm{CR}+\mathrm{CP}}.
\label{eq:citation_f1}
\end{equation}

\paragraph{C \quad Experiments and Analysis}

\paragraph{C.1 \quad Baselines.}

\begin{itemize}
    \item AutoSurvey adopts a two-stage generation strategy: it first retrieves relevant literature to construct a detailed outline, then employs multiple LLMs in parallel to generate individual chapters, and finally integrates these chapters into a coherent survey article.
    \item SurveyForge first generates an outline by analyzing the logical structure of human-written outlines and referring to the retrieved domain-related articles. It then automatically generates and refines the article content using an academic navigation agent that retrieves high-quality papers from memory.
    \item SurveyX decomposes the survey writing process into two stages: a preparation stage and a generation stage. By incorporating online reference retrieval, a preprocessing method called AttributeTree, and a refinement procedure, SurveyX significantly improves the efficiency of survey writing.
\end{itemize}

\paragraph{C.2 \quad Experimental Results.}

\begin{figure}[ht]
\centering
\includegraphics[width=1\columnwidth]{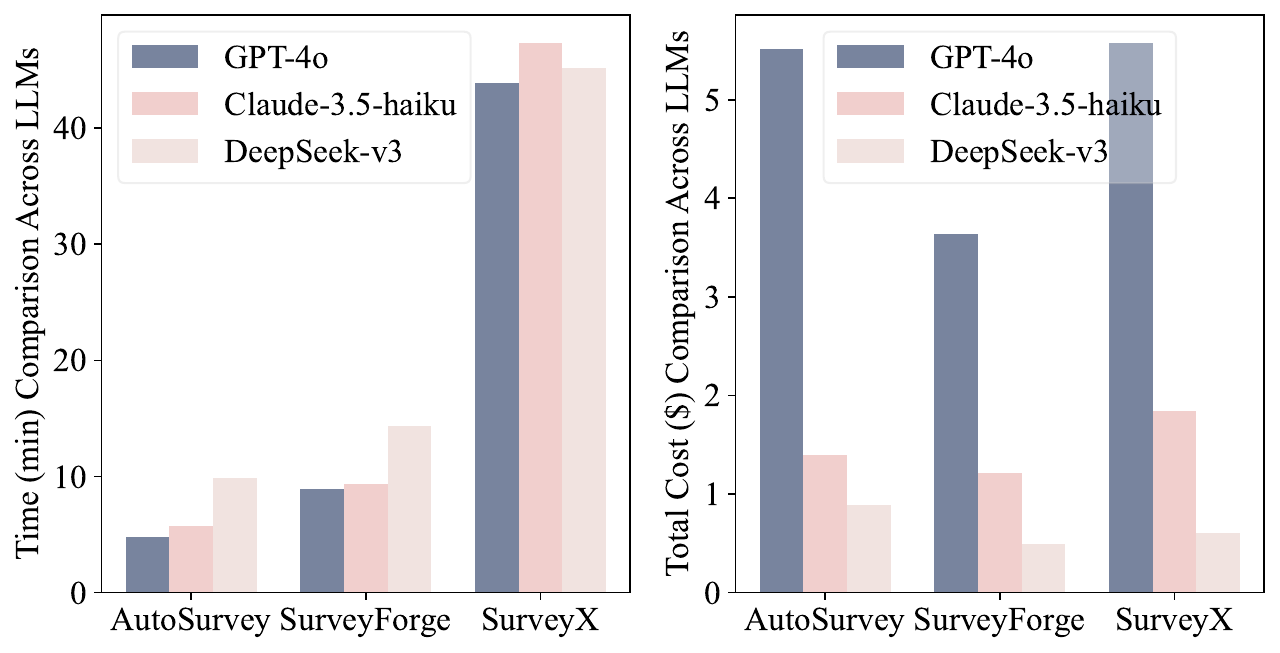} 
\caption{Comparison of generation time and API cost across survey generation methods and backbone LLMs.}
\label{defect}
\end{figure}

As shown in Figure~\ref{defect}, SurveyX requires substantially more generation time than AutoSurvey and SurveyForge across all backbone LLMs, taking approximately 44--48 minutes compared with 5--14 minutes for the other two methods. SurveyForge introduces only a moderate increase in runtime over AutoSurvey while consistently incurring the lowest API cost. Across all three methods, the overall cost is largely determined by the pricing of the backbone LLM, with GPT-4o being the most expensive, followed by Claude-3.5-Haiku and DeepSeek-v3. Overall, these results reveal a clear efficiency trade-off: SurveyX incurs the highest latency, whereas SurveyForge provides a more favorable balance between generation time and API expenditure.

\paragraph{C.3 \quad Human Evaluation and Inter-Evaluator Agreement Analysis}
To further assess the agreement between the multi-LLM ensemble evaluation and human judgments, we conduct a human evaluation on 40 surveys generated by AutoSurvey, SurveyX, and SurveyForge with GPT-4o as the backbone model. The selected surveys cover a diverse range of research topics. Three evaluators, comprising two doctoral students and an experienced researcher, independently score each survey according to the same 5-point academic value rubric used in the LLM-based evaluation. The final score for each metric is obtained by averaging the ratings of the three evaluators.

Table~\ref{tab:human} presents the human evaluation results. SurveyX achieves the highest average scores across all three academic value dimensions, with scores of 3.81 for informational value, 3.52 for scholarly communication value, and 3.90 for research guidance value. This overall performance pattern is consistent with the results obtained from the multi-LLM ensemble evaluation.

Both evaluation approaches consistently indicate that SurveyX achieves the best overall performance. These findings indicate that the principal conclusions derived from the multi-LLM ensemble evaluation are broadly aligned with human judgments, providing additional support for its reliability in assessing the academic value of generated surveys.

\paragraph{C.4 \quad Evidence Validation and Benchmark-Guided Improvement.}
For each selected survey, we randomly sample ten claim--evidence units from its main body, resulting in 400 units in total. Each unit consists of a factual claim and the complete set of references cited in support of that claim. The sampled units are manually examined from three perspectives. Factual Correctness assesses whether the factual content of a claim is consistent with the information reported in the cited literature. Claim--Evidence Alignment examines whether the cited references directly support the corresponding claim, rather than being merely relevant to the same general topic. Citation Reliability verifies whether the cited publications exist, whether their bibliographic information is correct, and whether their findings are accurately attributed in the generated survey. Each claim--evidence unit is independently assessed by the same three human evaluators, comprising two doctoral students and one experienced researcher. The preliminary label for each criterion is determined by majority voting. For units with inconsistent judgments, the evaluators jointly re-examine the claim, the relevant passages in the cited papers, and the predefined validation criteria to identify the source of disagreement. After discussion, a consensus label is assigned. For each criterion, the validation rate is calculated as the proportion of units receiving a positive final label among all examined units. The validation results are reported in Table~\ref{tab:evidence}. Among the 400 examined units, 376 are judged to be factually correct, 364 exhibit direct claim--evidence alignment, and 372 contain reliable citations, corresponding to validation rates of 94\%, 91\%, and 93\%, respectively. These results indicate that surveys receiving high academic value scores generally exhibit strong evidence integrity.

For each low-scoring survey, we use the metric-level diagnostic feedback from DeepSurvey-Bench to identify the content requiring revision. Then, the baseline method used to generate each scientific survey performs revisions solely based on the diagnostic feedback. During refinement, the survey topic, overall scope, outline structure, and reference set are kept unchanged. The model is not permitted to introduce new references or remove existing references, ensuring that improvements result from better organization, synthesis, comparison, and critical analysis of the existing literature rather than from additional literature retrieval. Each survey undergoes a single refinement iteration to avoid repeated optimization toward the evaluator. The original and refined surveys are evaluated using the same multi-LLM ensemble and 5-point rubric. As shown in Figure~\ref{fig:optimization}, refinement improves all seven academic value metrics, demonstrating the effectiveness of the diagnostic feedback provided by DeepSurvey-Bench.

\paragraph{D \quad Prompts and Scoring Rubrics.}
\paragraph{D.1 \quad Annotation Prompts and Scoring Rubrics}

The annotation prompt and scoring criteria are presented in Figure~\ref{fig: humanpkk}. Since the evaluation dimensions and criteria are identical to those described in Appendix~D.2, we provide only one representative example for illustration.

\begin{figure*}[!ht] 
\vspace{-1em}
\begin{AIbox}{Human Annotation: The Objective Clarity of Information Value}
{\color{black}\bf Role Description:} \\
We provide you with (1) an excerpt containing the academic-value related content extracted from a real scientific survey, and (2) the full survey paper in PDF form for reference. Based on these two sources, and strictly following our provided scoring rubric (1–5), please evaluate the academic value of this survey and assign a final score.
\tcbline
\textbf{Evaluation Dimensions:} \\
- **Research Objective Clarity**: Evaluate whether the research objective is specific, clear, and closely aligned with the core issues in the field.

- **Background and Motivation**: Evaluate whether the background and motivation are sufficiently explained, especially how they support the research objective.

- **Practical Significance and Guidance Value**: Evaluate whether the research objective demonstrates clear academic value and practical guidance for the field.
\tcbline
\textbf{Scoring Criteria:} 
\begin{lstlisting}[style=prompt]
{
    - **5 points**: The research objective is clear, specific, and the background and motivation are clearly explained. The objective has significant academic and practical value. For example, the objective is closely tied to the core issues of the field, and the review provides a thorough analysis of the current state and challenges.
    
    - **4 points**: The research objective is clear, but the background or motivation may be somewhat brief. The objective has noticeable academic or practical value, and it mostly guides the research direction clearly.
    
    - **3 points**: The research objective is present, but the background and motivation lack depth. The academic and practical value of the objective is not fully explained, and the objective is somewhat vague or lacks clear direction.
    
    - **2 points**: The objective is unclear, and the background and motivation are inadequately explained. The academic value of the objective is not clear, and some parts of the objective are vague or repetitive.
    
    - **1 point**: The review does not present a clear research objective, lacks background and motivation, and the research direction is not specified. The objective is almost nonexistent.
}
\end{lstlisting}
\tcbline
\textbf{Output Requirements:}\\
- A final academic-value score (1–5) and a brief justification explaining the key reasons for your score.

\end{AIbox} 
\caption{A prompt and score rubric of Human Annotation}
\label{fig: humanpkk}
\vspace{-1em}
\end{figure*}

\paragraph{D.2 \quad Evaluation Prompts and Scoring Rubrics}

The following figures present the prompts and scoring rubrics adopted for all LLM-based evaluations. Figures~8--11 present the evaluation prompts and 5-point scoring rubrics for the four content quality metrics: \textbf{Coverage}, \textbf{Structure}, \textbf{Relevance}, and \textbf{Language Fluency}. Figures~12--14 provide those for the three outline quality metrics: \textbf{Guidance for Content Generation}, \textbf{Hierarchical Clarity}, and \textbf{Logical Coherence}. Figures~15--21 present the prompts and rubrics for the seven fine-grained academic value metrics, organized into three dimensions: \textbf{Informational Value}, comprising \textbf{Objective Clarity}, \textbf{Classification--Evolution Coherence}, and \textbf{Dataset \& Metric Coverage}; \textbf{Scholarly Communication Value}, comprising \textbf{In-depth Comparison} and \textbf{Critical Analysis}; and \textbf{Research Guidance Value}, comprising \textbf{Research Gap} and \textbf{Future Work}. Each prompt specifies the evaluation objective, assessment instructions, and score definitions from 1 to 5, thereby standardizing the evaluation procedure across generated surveys.

\begin{figure*}[!ht] 
\vspace{-1em}
\begin{AIbox}{Coverage}
{\color{black}\bf Task Description:} \\
"coverage": Here is an academic survey about the topic [topic]:

---

[content]

---

<instruction>

Please evaluate this survey about the topic [topic] based on the criteria above provided below, and give a score from 1 to 5 according to the score description:

---

\tcbline
\textbf{Evaluation Description:} \\
Coverage: Coverage assesses the extent to which the survey encapsulates all relevant aspects of the topic, ensuring comprehensive discussion on both central and peripheral topics.

\tcbline
\textbf{Scoring Criteria:} 
\begin{lstlisting}[style=prompt]
{
    Score 1 Description: The survey has very limited coverage, only touching on a small portion of the topic and lacking discussion on key areas.

    Score 2 Description: The survey covers some parts of the topic but has noticeable omissions, with significant areas either underrepresented or missing.

    Score 3 Description: The survey is generally comprehensive in coverage but still misses a few key points that are not fully discussed.

    Score 4 Description: The survey covers most key areas of the topic comprehensively, with only very minor topics left out.

    Score 5 Description: The survey comprehensively covers all key and peripheral topics, providing detailed discussions and extensive information.
}
\end{lstlisting}
\tcbline
\textbf{Output Requirements:}\\
Return the score without any other information.
\end{AIbox} 
\caption{The prompt and score rubric of Coverage metric}
\label{fig: p1}
\vspace{-1em}
\end{figure*}

\begin{figure*}[!ht] 
\vspace{-1em}
\begin{AIbox}{Structure}
{\color{black}\bf Task Description:} \\
"structure": Here is an academic survey about the topic [topic]:

---

[content]

---

<instruction>

Please evaluate this survey about the topic [topic] based on the criteria above provided below, and give a score from 1 to 5 according to the score description:

---

\tcbline
\textbf{Evaluation Description:} \\
Structure: Structure evaluates the logical organization and coherence of sections and subsections, ensuring that they are logically connected.

\tcbline
\textbf{Scoring Criteria:} 
\begin{lstlisting}[style=prompt]
{
    Score 1 Description: The survey lacks logic, with no clear connections between sections, making it difficult to understand the overall framework.

    Score 2 Description: The survey has weak logical flow with some content arranged in a disordered or unreasonable manner.

    Score 3 Description: The survey has a generally reasonable logical structure, with most content arranged orderly, though some links and transitions could be improved such as repeated subsections.

    Score 4 Description: The survey has good logical consistency, with content well arranged and natural transitions, only slightly rigid in a few parts.

    Score 5 Description: The survey is tightly structured and logically clear, with all sections and content arranged most reasonably, and transitions between adajecent sections smooth without redundancy.
}
\end{lstlisting}
\tcbline
\textbf{Output Requirements:}\\
Return the score without any other information.
\end{AIbox} 
\caption{The prompt and score rubric of Structure metric}
\label{fig: p1}
\vspace{-1em}
\end{figure*}

\begin{figure*}[!ht] 
\vspace{-1em}
\begin{AIbox}{Relevance}
{\color{black}\bf Task Description:} \\
"relevance": Here is an academic survey about the topic [topic]:

---

[content]

---

<instruction>

Please evaluate this survey about the topic [topic] based on the criteria above provided below, and give a score from 1 to 5 according to the score description:

---

\tcbline
\textbf{Evaluation Description:} \\
Relevance: Relevance measures how well the content of the survey aligns with the research topic and maintain a clear focus.

\tcbline
\textbf{Scoring Criteria:} 
\begin{lstlisting}[style=prompt]
{
Score 1 Description: The  content is outdated or unrelated to the field it purports to review, offering no alignment with the topic.

Score 2 Description: The survey is somewhat on topic but with several digressions; the core subject is evident but not consistently adhered to.

Score 3 Description: The survey is generally on topic, despite a few unrelated details.

Score 4 Description: The survey is mostly on topic and focused; the narrative has a consistent relevance to the core subject with infrequent digressions.

Score 5 Description: The survey is exceptionally focused and entirely on topic; the article is tightly centered on the subject, with every piece of information contributing to a comprehensive understanding of the topic.
}
\end{lstlisting}
\tcbline
\textbf{Output Requirements:}\\
Return the score without any other information.
\end{AIbox} 
\caption{The prompt and score rubric of Relevance metric}
\label{fig: p1}
\vspace{-1em}
\end{figure*}

\begin{figure*}[!ht] 
\vspace{-1em}
\begin{AIbox}{Language Fluency}
{\color{black}\bf Task Description:} \\
"language fluency": Here is an academic survey about the topic [topic]:

---

[content]

---

<instruction>

Please evaluate this survey about the topic [topic] based on the criteria above provided below, and give a score from 1 to 5 according to the score description:

---

\tcbline
\textbf{Evaluation Description:} \\
Language Fluency: Language Fluency assesses the academic formality, clarity, and correctness of the writing, including grammar, terminology, and tone.

\tcbline
\textbf{Scoring Criteria:} 
\begin{lstlisting}[style=prompt]
{
        Score 1 Description: The language is highly informal, contains frequent grammatical errors, imprecise terminology, and numerous colloquial expressions. The writing lacks academic tone and professionalism.
        
        Score 2 Description: The writing style is somewhat informal, with several grammatical errors or ambiguous expressions. Academic terminology is inconsistently used.
        
        Score 3 Description: The language is mostly formal and generally clear, with only occasional minor grammatical issues or slightly informal phrasing.
        
        Score 4 Description: The language is clear, formal, and mostly error-free, with only rare lapses in academic tone or minor imprecisions.
        
        Score 5 Description: The writing is exemplary in academic formality and clarity, using precise terminology throughout, flawless grammar, and a consistently scholarly tone.
}
\end{lstlisting}
\tcbline
\textbf{Output Requirements:}\\
Return the score without any other information.
\end{AIbox} 
\caption{The prompt and score rubric of Language metric}
\label{fig: p1}
\vspace{-1em}
\end{figure*}

\begin{figure*}[!ht] 
\vspace{-1em}
\begin{AIbox}{Guidance for Content Generation}
{\color{black}\bf Input Standard:} \\
Here is an academic survey outline about the topic [topic]:

--

[content]

--

\tcbline
\textbf{Evaluation Description:} \\
Criteria Description Guidance for Content Generation: Does the outline effectively guide content generation, ensuring comprehensive coverage of the topic? 

\tcbline
\textbf{Scoring Criteria:} 
\begin{lstlisting}[style=prompt]
{
    Score 1 Description The outline fails to guide content generation, omitting significant aspects of the topic or providing insufficient direction. 
    
    Score 2 Description The outline provides limited guidance, covering some key areas but lacking depth or completeness in addressing the topic. 
    
    Score 3 Description The outline provides moderate guidance for content generation, addressing most key areas but leaving some gaps or ambiguities
    . 
    Score 4 Description The outline effectively guides content generation, covering all significant aspects with clear direction, though minor refinements could enhance comprehensiveness. 
    
    Score 5 Description The outline is exemplary in guiding content generation, thoroughly addressing all aspects of the topic with clear, detailed direction and no significant gaps.
}
\end{lstlisting}
\tcbline
\textbf{Output Requirements:}\\
Return the score without any other information.

\end{AIbox} 
\caption{The prompt and score rubric of Guidance metric}
\label{fig: humanp1}
\vspace{-1em}
\end{figure*}

\begin{figure*}[!ht] 
\vspace{-1em}
\begin{AIbox}{Hierarchical Clarity}
{\color{black}\bf Input Standard:} \\
Here is an academic survey outline about the topic [topic]:

--

[content]

--

\tcbline
\textbf{Evaluation Description:} \\
Criteria Description Hierarchical Clarity: Does the outline clearly define a hierarchy of topics and subtopics, with a logical, diverse structure that is easy to understand? 

\tcbline
\textbf{Scoring Criteria:} 
\begin{lstlisting}[style=prompt]
{
    Score 1 Description The outline exhibits no discernible hierarchical structure. Topics and subtopics are jumbled together without logical separation or clear levels, making it nearly impossible to follow or identify any organization. 

    Score 2 Description The outline attempts to establish a hierarchy but fails to maintain logical consistency. Main topics and subtopics are frequently misclassified, and the structure is overly rigid or disjointed. Subtopics may be missing, misplaced, or redundant, making it hard to grasp the intent of the structure. 

    Score 3 Description The outline has a recognizable hierarchical structure but lacks diversity in organization style. While main topics are somewhat clear, subtopics occasionally overlap, are misaligned, or follow a repetitive format. This restricts flexibility and introduces mild confusion in certain areas. 

    Score 4 Description The outline displays a clear, logical, and diverse hierarchical structure. Main topics are distinct, and subtopics are properly nested. While most elements are well-placed, there may be minor redundancies or opportunities to introduce more diverse formats for subtopics. Slight adjustments could achieve better precision and variety in style. 

    Score 5 Description The outline showcases an exceptional, flawless hierarchical structure. Each main topic is distinct, and subtopics are logically nested with absolute clarity and stylistic diversity. The outline demonstrates flexibility in structure and organization, adapting its style where appropriate for the content and logic. No further refinement is necessary.

}
\end{lstlisting}
\tcbline
\textbf{Output Requirements:}\\
Return the score without any other information.

\end{AIbox} 
\caption{The prompt and score rubric of Hierarchical Clarity metric}
\label{fig: humanp1}
\vspace{-1em}
\end{figure*}

\begin{figure*}[!ht] 
\vspace{-1em}
\begin{AIbox}{Logical Coherence}
{\color{black}\bf Input Standard:} \\
Here is an academic survey outline about the topic [topic]:

--

[content]

--

\tcbline
\textbf{Evaluation Description:} \\
Criteria Description Logical Coherence: Does the outline logically organize topics and subtopics, ensuring a smooth and natural flow of ideas with clear logical transitions?

\tcbline
\textbf{Scoring Criteria:} 
\begin{lstlisting}[style=prompt]
{
    Score 1 Description The outline is highly disjointed and incoherent. Topics and subtopics appear in a random, unordered manner, with no logical flow or sense of progression. Major conceptual gaps and illogical jumps are present throughout the structure. 
    
    Score 2 Description The outline shows some attempt at logical organization, but it contains frequent inconsistencies, abrupt shifts, or logical missteps. Topics and subtopics are misaligned or lack proper transitions, making the reader work hard to follow the structure. 
    
    Score 3 Description The outline demonstrates a basic level of logical coherence. Most topics follow a general sequence, but some sections feel forced, with weak or unclear transitions. There are small jumps in logic, causing slight confusion or loss of flow at certain points. 
    
    Score 4 Description The outline exhibits a strong sense of logical flow, with ideas presented in a mostly smooth and connected manner. Transitions between topics and subtopics are clear, but a few minor adjustments could make the flow more seamless or natural. The logic is sound, but room for refinement exists. 
    
    Score 5 Description The outline achieves exceptional logical coherence. Each topic and subtopic follows a deliberate, thoughtful progression, with clear, natural, and intuitive transitions. The reader experiences a seamless flow of ideas, and no adjustments are required to improve logical consistency or flow.
}
\end{lstlisting}
\tcbline
\textbf{Output Requirements:}\\
Return the score without any other information.

\end{AIbox} 
\caption{The prompt and score rubric of Logical Coherence metric}
\label{fig: humanp1}
\vspace{-1em}
\end{figure*}


\begin{figure*}[!ht] 
\vspace{-1em}
\begin{AIbox}{Information Value: Objective Clarity}
{\color{black}\bf Role Description:} \\
You are now acting as an **experienced literature review evaluator** with years of academic review experience. You are proficient in evaluating the clarity of research objectives, the articulation of background and motivation, and the clarity of research direction in academic papers. You will evaluate the **Abstract** and **Introduction** sections of the paper in detail.
\tcbline
\textbf{Evaluation Dimensions:} \\
- **Research Objective Clarity**: Evaluate whether the research objective is specific, clear, and closely aligned with the core issues in the field.
- **Background and Motivation**: Evaluate whether the background and motivation are sufficiently explained, especially how they support the research objective.

- **Practical Significance and Guidance Value**: Evaluate whether the research objective demonstrates clear academic value and practical guidance for the field.
\tcbline
\textbf{Scoring Criteria:} 
\begin{lstlisting}[style=prompt]
{
    - **5 points**: The research objective is clear, specific, and the background and motivation are clearly explained. The objective has significant academic and practical value. For example, the objective is closely tied to the core issues of the field, and the review provides a thorough analysis of the current state and challenges.
    
    - **4 points**: The research objective is clear, but the background or motivation may be somewhat brief. The objective has noticeable academic or practical value, and it mostly guides the research direction clearly.
    
    - **3 points**: The research objective is present, but the background and motivation lack depth. The academic and practical value of the objective is not fully explained, and the objective is somewhat vague or lacks clear direction.
    
    - **2 points**: The objective is unclear, and the background and motivation are inadequately explained. The academic value of the objective is not clear, and some parts of the objective are vague or repetitive.
    
    - **1 point**: The review does not present a clear research objective, lacks background and motivation, and the research direction is not specified. The objective is almost nonexistent.
}
\end{lstlisting}
\tcbline
\textbf{Output Requirements:}\\
- Please **first provide the score** for this section (1-5 points).

- **Then provide a detailed explanation** of why you assigned this score, and specifically mention which parts of the paper (including chapters and sentences) support your scoring.

- Please ensure that the score is entirely consistent with the content, and make a reasonable judgment based on the actual content of the paper.
\end{AIbox} 
\caption{The prompt and score rubric of Objective Clarity metric}
\label{fig: p1}
\vspace{-1em}
\end{figure*}

\begin{figure*}[!ht] 
\vspace{-1em}
\begin{AIbox}{Information Value: Classification-Evolution Coherence}
{\color{black}\bf Role Description:} \\
You are now acting as a **senior literature review evaluator** with many years of academic review experience. You are proficient in evaluating the clarity and reasonableness of the method classification system and the evolution of the technological progression in literature reviews. You will evaluate the **Method** and/or **Related Work** sections of the paper. If the paper does not explicitly use headings such as "Method" or "Related Work", focus on the content after the **Introduction** and before the **Experiments/Evaluation** sections for a detailed evaluation.
\tcbline
\textbf{Evaluation Dimensions:} \\
- **Method Classification Clarity**: Evaluate whether the method classification is clear and reasonable and whether it reflects the technological development path in the field.\\
- **Evolution of Methodology**: Evaluate whether the evolution process of methods is systematically presented and whether the technological or methodological trends are shown.
\tcbline
\textbf{Scoring Criteria:}
\begin{lstlisting}[style=prompt]
{
    - **5 points**: The method classification is completely clear, and the evolution process is systematically presented, well revealing the technological advancements and field development trends. Each category is clearly defined with inherent connections, and the evolutionary direction of methods is clear and innovative.
    
    - **4 points**: The method classification is relatively clear, and the evolution process is somewhat presented, but the connections between some methods are unclear, and some evolutionary stages are not fully explained. Overall, it reflects the technological development of the field.
    
    - **3 points**: The method classification is somewhat vague, and the evolution process is partially clear, but lacks a detailed analysis of the inheritance between methods. Some evolutionary directions are unclear.
    
    - **2 points**: The method classification is unclear, the evolution process is not well-defined, and there is no analysis of the relationships between methods. It is difficult to clearly present the technological progress of the field.
    
    - **1 point**: The method classification is chaotic, and the evolution process is almost unrecognizable. There is no mention of technological progress or the relationships between methods.
}
\end{lstlisting}
\tcbline
\textbf{Output Requirements:}\\
- Please **first provide the score** for this section (1-5 points).

- **Then provide a detailed explanation** of why you assigned this score, and specifically mention which parts of the paper (including chapters and sentences) support your scoring.

- Please ensure that the score is entirely consistent with the content, and make a reasonable judgment based on the actual content of the paper.
\end{AIbox} 
\caption{The prompt and score rubric of Classification-Evolution Coherence metric}
\label{fig: p2}
\vspace{-1em}
\end{figure*}

\begin{figure*}[!ht] 
\vspace{-1em}
\begin{AIbox}{Information Value: Dataset \& Metric Coverage}
{\color{black}\bf Role Description:} \\
You are now acting as a **senior literature review evaluator** with many years of academic review experience. You are proficient in evaluating the coverage of datasets and the applicability of evaluation metrics in the literature review. You will carefully evaluate the **Data**, **Evaluation**, and/or **Experiments** sections of the paper.
\tcbline
\textbf{Evaluation Dimensions:} \\
- **Diversity of Datasets and Metrics**: Evaluate whether the review covers a variety of datasets and evaluation metrics, and whether it includes important datasets and metrics in the field.

- **Rationality of Datasets and Metrics**: Evaluate whether the choice of datasets is reasonable and sufficiently supports the research objective, and whether the evaluation metrics are academically sound and practically meaningful.
\tcbline
\textbf{Scoring Criteria:}
\begin{lstlisting}[style=prompt]
{
    - **5 points**: The review comprehensively covers multiple datasets and evaluation metrics, providing detailed descriptions of each dataset's scale, application scenario, and labeling method. The choice and use of evaluation metrics are highly targeted and reasonable, covering the key dimensions of the field.
    
    - **4 points**: The review includes multiple datasets and evaluation metrics, and the description of each dataset is fairly detailed. The choice of evaluation metrics is generally reasonable, but some aspects of the dataset's application scenarios or the use of metrics may not be fully explained.
    
    - **3 points**: The review covers a limited set of datasets and evaluation metrics, and the descriptions lack detail. The choice of metrics does not fully reflect key dimensions of the field, and the characteristics of datasets are not sufficiently explained.
    
    - **2 points**: The review includes few datasets or evaluation metrics, and the descriptions are not clear or detailed. There is a lack of analysis of the rationale behind the choices, and some datasets or metrics are not described in detail.
    
    - **1 point**: No datasets or evaluation metrics are mentioned, or the descriptions are extremely simple and not practical.
}
\end{lstlisting}
\tcbline
\textbf{Output Requirements:}\\
- Please **first provide the score** (1-5 points) for this section.

- **Then provide a detailed explanation**, explaining why you assigned this score, and specifically mention which parts of the paper (including chapters and sentences) support your scoring.

- Please ensure that the score is entirely consistent with the content, and make a reasonable judgment based on the actual content of the paper.
\end{AIbox} 
\caption{The prompt and score rubric of Dataset \& Metric Coverage metric}
\label{fig: p3}
\vspace{-1em}
\end{figure*}

\begin{figure*}[!ht] 
\vspace{-1em}
\begin{AIbox}{Scholarly communication value: In-depth Comparison}
{\color{black}\bf Role Description:} \\
You are now acting as a **senior literature review evaluator** with many years of academic review experience. You are proficient in evaluating the comparison of different research methods and the analysis of their advantages, disadvantages, similarities, and distinctions in literature review papers. You will carefully evaluate the **Method** and/or **Related Work** sections of the paper. If the paper does not explicitly use "Method" or "Related Work" as section titles, focus on the content after the **Introduction** and before the **Experiments/Evaluation** sections for a detailed evaluation.
\tcbline
\textbf{Evaluation Dimensions:} \\
Evaluate the **clarity, rigor, and depth** of the review’s **comparison of different research methods**. This evaluation focuses on whether the paper:

- systematically compares methods across multiple dimensions
- clearly describes **advantages and disadvantages**
- identifies **commonalities and distinctions**
- explains differences in terms of **architecture, objectives, or assumptions**
- avoids superficial or fragmented listing of methods

This dimension emphasizes **objective and structured comparison**, rather than subjective commentary.
\tcbline
\textbf{Scoring Criteria:}
\begin{lstlisting}[style=prompt]
{
    - **5 points**:  
    The review presents a **systematic, well-structured, and detailed comparison** of multiple methods, clearly summarizing their advantages, disadvantages, commonalities, and distinctions across multiple meaningful dimensions (e.g., modeling perspective, data dependency, learning strategy, application scenario). The comparison is technically grounded and reflects a comprehensive understanding of the research landscape.

    - **4 points**:  
    The review provides a **clear comparison** of major advantages and  disadvantages of the methods and identifies their similarities and differences, but some comparison dimensions are not fully elaborated, or certain aspects of the comparison remain at a relatively high level.

    - **3 points**:  
    The review mentions the **pros and cons** or **differences** between methods, but the comparison is **partially fragmented or superficial**, lacking systematic structure or sufficient technical depth in contrasting the methods.

    - **2 points**:  
    The review mainly **lists the characteristics or outcomes** of different methods, with limited explicit comparison. Advantages and disadvantages are mentioned in isolation, and **the relationships among methods are not clearly contrasted**.

    - **1 point**:  
    The review does not provide meaningful comparison. Methods are described **independently**, with no discussion of similarities, differences, or advantages and disadvantages.
}
\end{lstlisting}
\tcbline
\textbf{Output Requirements:}\\
- Please **first provide the score** (1-5 points) for this section.

- **Then provide a detailed explanation**, explaining why you assigned this score, and specifically mention **which sections and sentences** in the paper support your scoring.

- Please ensure that the score is **entirely consistent with the content**, and make a reasonable judgment based on the actual content of the paper.
\end{AIbox} 
\caption{The prompt and score rubric of In-depth Comparison metric}
\label{fig: p4}
\vspace{-1em}
\end{figure*}

\begin{figure*}[!ht] 
\vspace{-1em}
\begin{AIbox}{Scholarly communication value: Critical Analysis}
{\color{black}\bf Role Description:} \\
You are now acting as a **senior literature review evaluator** with many years of academic review experience. You are proficient in evaluating the **critical analysis, interpretation, and reflective commentary** provided in literature review papers. You will carefully evaluate the **Method** and/or **Related Work** sections of the paper. If the paper does not explicitly use "Method" or "Related Work" as section titles, focus on the content after the **Introduction** and before the **Experiments/Evaluation** sections for a detailed evaluation.
\tcbline
\textbf{Evaluation Dimensions:} \\
Evaluate the **depth, reasoning, and insightfulness** of the review’s **critical analysis of different methods**. This evaluation focuses on whether the paper:

- explains the **fundamental causes** of differences between methods
- analyzes **design trade-offs, assumptions, and limitations**
- synthesizes relationships across research lines
- provides **technically grounded explanatory commentary**
- extends beyond descriptive summary to offer **interpretive insights**

This dimension emphasizes **analytical reasoning and reflective interpretation**, not merely reporting or summarization.
\tcbline
\textbf{Scoring Criteria:}
\begin{lstlisting}[style=prompt]
{
    - **5 points**:  
    The review provides **deep, well-reasoned, and technically grounded critical analysis**, clearly explaining the **underlying mechanisms, design trade-offs, and fundamental causes** of methodological differences. It synthesizes connections across research directions and offers **insightful, evidence-based personal commentary** that meaningfully interprets the development trends and limitations of existing work.

    - **4 points**:  
    The review offers **meaningful analytical interpretation** of method differences and provides reasonable explanations for some underlying causes, but the depth of analysis is **uneven across methods**, or some arguments remain partially underdeveloped.

    - **3 points**:  
    The review includes **basic analytical comments** or evaluative statements, but the analysis remains **relatively shallow**, focusing more on descriptive remarks than on rigorous technical reasoning. Explanations of fundamental causes are **limited or implicit**.

    - **2 points**:  
    The review provides only **brief or generic evaluative comments** without explaining why such differences or limitations arise. Arguments lack analytical depth and do not meaningfully interpret relationships between methods.

    - **1 point**:  
    The review lacks **critical analysis**. The paper only presents methods descriptively, with no interpretive commentary, reasoning, or reflective insight.
}
\end{lstlisting}
\tcbline
\textbf{Output Requirements:}\\
- Please **first provide the score** (1-5 points) for this section.

- **Then provide a detailed explanation**, explaining why you assigned this score, and specifically mention **which sections and sentences** in the paper support your scoring.

- Please ensure that the score is **entirely consistent with the content**, and make a reasonable judgment based on the actual content of the paper.
\end{AIbox} 
\caption{The prompt and score rubric of Critical Analysis metric}
\label{fig: p5}
\vspace{-1em}
\end{figure*}

\begin{figure*}[!ht] 
\vspace{-1em}
\begin{AIbox}{Research guidance value: Research Gaps}
{\color{black}\bf Role Description:} \\
You are now acting as a **senior literature review evaluator** with many years of academic review experience. You are proficient in evaluating the identification and analysis of research gaps in literature review papers. You will carefully evaluate the **Gap/Future Work** section of the paper.
\tcbline
\textbf{Evaluation Dimensions:} \\
Evaluate whether the review systematically identifies, analyzes, and explains the key issues and shortcomings that need to be addressed in the future of the research field, based on the current achievements. This evaluation focuses not only on whether the "unknowns" are pointed out but also on the **depth of analysis** regarding why these issues are important and what impact they have.
\tcbline
\textbf{Scoring Criteria:}
\begin{lstlisting}[style=prompt]
{
    - **5 points**:  
    Based on the review content, the **major research gaps** are comprehensively identified and deeply analyzed, covering **data**, **methods**, and other dimensions. The analysis is detailed and discusses the **potential impact** of each gap on the development of the field.

    - **4 points**:  
    The review points out several research gaps, but the analysis is somewhat **brief** and does not delve deeply into the **impact** or **background** of each gap. The gaps are identified in a comprehensive way, but the discussion is not fully developed.

    - **3 points**:  
    The review lists some research gaps but lacks in-depth analysis or discussion. Although some gaps are identified, their **impact** or **reasons** are not fully explored.

    - **2 points**:  
    The research gap analysis is **limited**, and the review does not fully discuss the identified gaps. The gaps are mentioned in passing but not explored in detail.

    - **1 point**:  
    The review **does not identify or discuss any research gaps**, and there is a lack of analysis of the **major issues** in the field.
}
\end{lstlisting}
\tcbline
\textbf{Output Requirements:}\\
- Please **first provide the score** (1-5 points) for this section.

- **Then provide a detailed explanation**, explaining why you assigned this score, and specifically mention **which sections and sentences** in the paper support your scoring.

- Please ensure that the score is **entirely consistent with the content**, and make a reasonable judgment based on the actual content of the paper.
\end{AIbox} 
\caption{The prompt and score rubric of Research Gaps metric}
\label{fig: p6}
\vspace{-1em}
\end{figure*}

\begin{figure*}[!ht] 
\vspace{-1em}
\begin{AIbox}{Research guidance value: Future Work}
{\color{black}\bf Role Description:} \\
You are now acting as a **senior literature review evaluator** with many years of academic review experience. You are proficient in evaluating the identification of future research directions and their innovative analysis in literature review papers. You will carefully evaluate the **Gap/Future Work** section of the paper.
\tcbline
\textbf{Evaluation Dimensions:} \\
Evaluate whether the paper proposes **forward-looking research directions** based on the **existing research gaps** or **real-world issues**, and whether it offers **new research topics** or **suggestions** in response to these gaps, aligning with real-world needs.
\tcbline
\textbf{Scoring Criteria:}
\begin{lstlisting}[style=prompt]
{
    - **5 points**:  
    The review tightly integrates the **key issues** and **research gaps** in the field, proposing **highly innovative** research directions that effectively address **real-world needs**. The review presents **specific and innovative research topics or suggestions** and provides a thorough analysis of their **academic and practical impact**, offering a **clear and actionable path** for future research.

    - **4 points**:  
    The review identifies several **forward-looking research directions** based on key issues and research gaps, addressing **real-world needs**, but the analysis of the **potential impact** and **innovation** is somewhat shallow. The directions are innovative, but the discussion is brief and does not fully explore the **causes** or **impacts** of the research gaps.

    - **3 points**:  
    The proposed research directions are **broad** and lack an in-depth discussion of their **forward-looking nature**. The review does not clearly explain how these directions address the **existing research gaps** or meet **real-world needs**. 

    - **2 points**:  
    The future research directions are unclear and lack an in-depth analysis of their **innovation**. The proposed directions are **traditional**, and the review does not clearly explain their **academic significance** or **practical value**. 

    - **1 point**:  
    The review does not propose any **future research directions**, nor does it discuss the **forward-looking nature** of the research. 
}
\end{lstlisting}
\tcbline
\textbf{Output Requirements:}\\
- Please **first provide the score** (1-5 points) for this section.

- **Then provide a detailed explanation**, explaining why you assigned this score, and specifically mention **which sections and sentences** in the paper support your scoring.

- Please ensure that the score is **entirely consistent with the content**, and make a reasonable judgment based on the actual content of the paper.
\end{AIbox} 
\caption{The prompt and score rubric of Future Work metric}
\label{fig: p7}
\vspace{-1em}
\end{figure*}


\end{document}